\documentclass[pdflatex,sn-basic]{sn-jnl}

\usepackage{comment}
\usepackage{wrapfig}
\usepackage{amsmath,amssymb} 
\usepackage{color}
\usepackage{soul}
\usepackage{url}
\usepackage{graphicx}
\usepackage{amsmath}
\usepackage{float}
\usepackage{microtype}
\usepackage{amssymb}
\usepackage{makecell}
\usepackage{array}
\usepackage{booktabs}
\usepackage{xcolor,colortbl}
\definecolor{Gray}{gray}{0.95}
\newcolumntype{a}{>{\columncolor{Gray}}p}
\definecolor{blue(pigment)}{rgb}{0., 0.1484, 0.6992}
\hypersetup{colorlinks,linkcolor=blue(pigment),citecolor=blue(pigment)}

\jyear{2021}%

\theoremstyle{thmstyleone}%
\newtheorem{theorem}{Theorem}
\theoremstyle{thmstyletwo}%

\theoremstyle{thmstylethree}%

\begin{document}

\title[3DVerifier: Efficient Robustness Verification for 3D Point Cloud Models]
{3DVerifier: Efficient Robustness Verification for 3D Point Cloud Models } 


\author[1]{\fnm{Ronghui} \sur{Mu}}

\author[2]{\fnm{Wenjie} \sur{Ruan}}
\author[1]{\fnm{Leandro S.} \sur{Marcolino}}
\author[1]{\fnm{Qiang} \sur{Ni}}
\affil[1]{\orgdiv{School of Computing \& Communication}, \orgname{Lancaster University}, \orgaddress{
\country{UK}}}

\affil[2]{\orgdiv{Department of Computer Science}, \orgname{University of Exeter}, \orgaddress{ 
\country{UK}}}


\abstract{3D point cloud models are widely applied in safety-critical scenes, which delivers an urgent need to obtain more solid proofs to verify the robustness of models. Existing verification method for point cloud model is time-expensive and computationally unattainable on large networks. Additionally, they cannot handle the complete PointNet model with joint alignment network (JANet) that contains multiplication layers, which effectively boosts the performance of 3D models. This motivates us to design a more efficient and general framework to verify various architectures of point cloud models. The key challenges in verifying the large-scale complete PointNet models are addressed as dealing with the cross-non-linearity operations in the multiplication layers and the high computational complexity of high-dimensional point cloud inputs and added layers. Thus, we propose an efficient verification framework, 3DVerifier, to tackle both challenges by adopting a linear relaxation function to bound the multiplication layer and combining forward and backward propagation to compute the certified bounds of the outputs of the point cloud models. Our comprehensive experiments demonstrate that 3DVerifier outperforms existing verification algorithms for 3D models in terms of both efficiency and accuracy. Notably, our approach achieves an orders-of-magnitude improvement in verification efficiency for the large network, and the obtained certified bounds are also significantly tighter than the state-of-the-art verifiers. We release our tool 3DVerifier via \url{https://github.com/TrustAI/3DVerifier} for use by the community.}

\keywords{Robustness Verification; 3D Point Cloud Models; Adversarial Attacks; Deep Neural Networks; PointNet}
\maketitle
\section{Introduction}

Recent years have witnessed increasing interest in 3D object detection, and Deep Neural Networks (DNNs) have also demonstrated their remarkable performance in this area~\citep{qi2017pointnet,qi2017pointnet++}. For 3D object detectors, the point clouds are utilized to represent 3D objects, which are usually the raw data gained from LIDARs and depth cameras. Such 3D deep learning models have been widely employed in multiple safety-critical applications such as motion planning ~\citep{varley2017shape}, virtual reality ~\citep{vr3d}, and autonomous driving ~\citep{chen2017multiview,liang2018deep}. However, extensive research has shown that DNNs are vulnerable to adversarial attacks, appearing as adding a small amount of non-random, ideally human-invisible, perturbations on the input will cause DNNs to make abominable predictions ~\citep{szegedy2014intriguing,carlini2017evaluating,jin2020does}. Therefore, there is an urgent need to address such safety concerns on DNNs caused by adversarial examples, especially on safety-critical 3D object detection scenarios.

Recently, most works on analyzing the robustness of 3D models mainly focus on adversarial attacks with an aim to reveal the model's vulnerabilities under different types of perturbations, such as adding or removing points, and shifting positions of points ~\citep{liu2019extending,yang2021adversarial}. In the meanwhile, adversarial defenses are also proposed to detect or prevent these adversarial attacks ~\citep{yang2021adversarial,zhou2019dup}. However, as \citet{tramer2020adaptive} and \citet{athalye2018obfuscated} indicated, {\em even though these defenses are effective for some attacks, they still can be broken by other stronger attacks}. Thereby, we need a more solid solution, ideally with {\it provable guarantees}, to verify whether the model is robust to {\em any} adversarial attacks within an allowed perturbation budget\footnote{In the community, we normally use a small predefined $l_p$-norm ball to quantify such perturbations, namely, within this small perturbing space the decision should remain the same from a perspective of a human observer.}. This technique is also generally regarded as verification on (local) adversarial robustness\footnote{For convenience, we use {\em robustness verification} or {\em verification} for short in this paper.} in the community. So far, various solutions have been proposed to tackle robustness verification, but they mostly focus on the image domain ~\citep{boopathy2018cnncert,singh2019abstract,tjeng2018evaluating,jin2022enhancing}. Verifying the adversarial robustness of 3D point cloud models, by contrast, is barely explored by the community. As far as we know, 3DCertify, proposed by \citet{lorenz2021robustness} is the first and also the {\em only} work to verify the robustness of 3D models. Although 3DCertify is very inspiring, it has yet completely resolved some key challenges on robustness verification for 3D models according to our empirical study. 

\begin{figure}
    \begin{center}
    \includegraphics[scale=0.3]{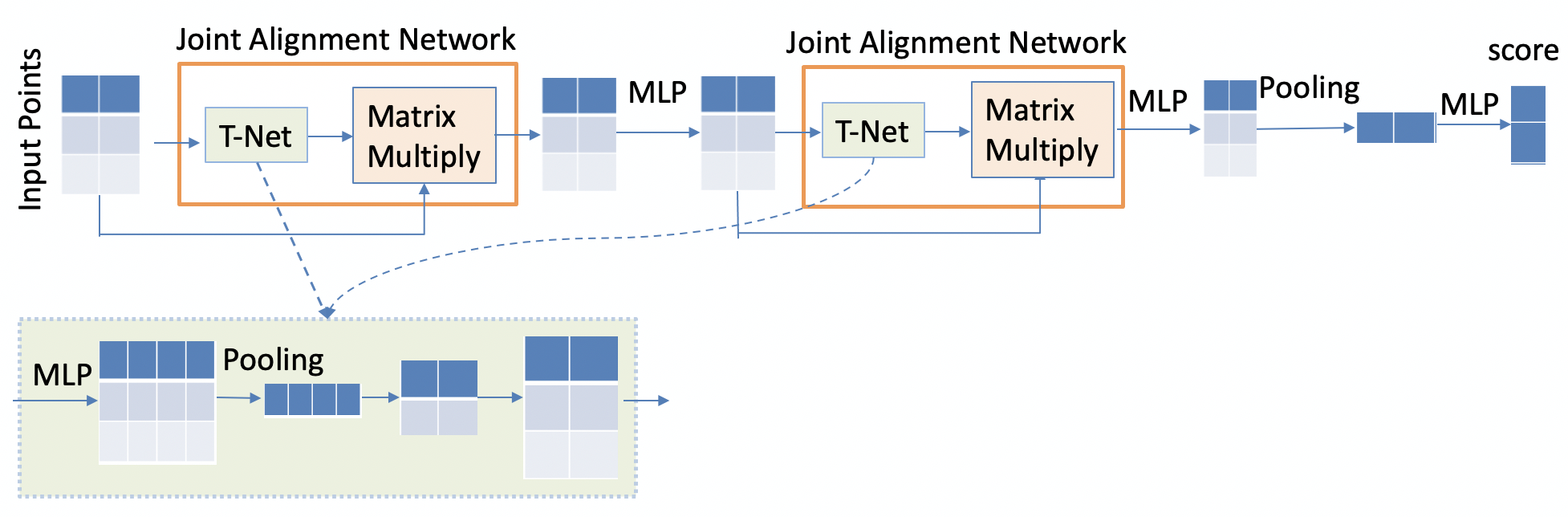}
    \end{center}
    \caption{{The abstract framework of the PonitNet with joint alignment network (JANet), where MLP stands for multi-layer perceptron.}}
    \label{fig:pointnet}
    \vspace{-3mm}
\end{figure}

Firstly, as the first verification tool for 3D models, 3DCertify is time-consuming and thus not computationally attainable on large neural networks. As 3DCertify is built upon DeepPoly ~\citep{singh2019abstract}, when directly applying the relaxation algorithm that is specifically designed for images on the high-dimensional point clouds, 
it will result in out-of-memory issues and cause the termination of the verification. Moreover, 3DCertify can only verify a simplified PointNet model without Joint Alignment Network (JANet) consisting of matrix multiplication operations.
Figure \ref{fig:pointnet} illustrates the abstract architecture of a complete PointNet ~\citep{qi2017pointnet}, one of the most widely used models on 3D object detection. As we can see, since the learnt representations are expected to be invariant to spatial transformations, JANet is the key enabler in PointNet to achieve this geometric invariant functionality by adopting the T-Net and matrix multiplications. Recent research also demonstrates that JANet is an essential for boosting the performance of PointNet ~\citep{qi2017pointnet} and thus widely applied in some safety-critical tasks  ~\citep{paigwar2019attentional,Aoki_2019_CVPR,chen2021individual}. Thirdly, 3DCertify can only work on $l_{\infty}$-norm metric, however, arguably, some researchers in the community regard other $l_p$-norm metrics such as $l_1$, $l_2$-norm metrics are equally (if not more) important in the study of adversarial robustness ~\citep{boopathy2018cnncert,weng2018fast}. Thus, a robustness verification tool that can work on a wide range of $l_p$-norm metrics is also worthy of a comprehensive exploration.

Thus, motivated by the aforementioned challenges yet to be resolved, this paper aims to design an {\em efficient} and {\em scalable} robustness verification tool that can handle a wide range of 3D models, including those with JANet structures under multiple $l_p$-norm metrics including $l_{\infty}$, $l_1$, and $l_2$-norm. We achieve the verification efficiently by adapting the efficient layer-by-layer certification framework used in ~\citep{weng2018fast,boopathy2018cnncert}. Considering that these verifiers are designed for images and cannot be applied in larger-scale 3D point cloud models, we introduce a novel relaxation function of global max pooling to make it applicable and efficient on PointNet. Moreover, the multiplication layers in the JANet structure involves two variables under perturbations, which brings the cross-non-linearity.
Due to the high dimensionality in 3D models, such cross-non-linearity results in significant computational overhead for computing a tight bound. Inspired by the recent advance of certification on transformers ~\citep{Shi2020Robustness}, we propose closed-form linear functions to bound the multiplication layer and combine forward and backward propagation to speed up the bound computation, which can be calculated in only $O(1)$ complexity. In summary, the main contributions of this paper are listed as below. 
\begin{itemize}
    \item  Our method can achieve an efficient verification. We design a relaxation algorithm to resolve the cross-non-linearity challenge by combining the forward and backwards propagation, enabling an efficient yet tight verification on matrix multiplications. 

    \item We design an efficient and scalable verification tool, 3DVerifier, with provable guarantees. It is a general framework that can verify the robustness for a wide range of 3D model architectures, especially it can work on complete and large-scale 3D models under $l_{\infty}$, $l_1$, and $l_2$-norm perturbations.
    
    \item 3DVerifier, as far as we know, is one of the very few works on 3D model verification, which is more advanced than the existing work, 3DCertify, in terms of efficiency, scalability and tightness of the certified bounds.
\end{itemize}

\section{Related Work}
\textbf{3D Deep Learning Models:} As the raw data of point clouds comes directly from real-world sensors, such as LIDARs and depth cameras, point clouds are widely used to represent 3D objects for the deep learning classification task. To deal with the unordered point clouds, \citet{qi2017pointnet} proposed the PointNet model, which utilised the global max-pooling layer to assemble features of points and the joint alignment network (JANet). Then, \citet{qi2017pointnet++} extended the PointNet by using the spacial neighbor graphs. 

\textbf{Adversarial Attacks on 3D Point Clouds Classification:} \citet{szegedy2014intriguing} first proposed the concepts of adversarial attacks and indicated that the neural networks are vulnerable to the well-crafted imperceptible perturbation. \citet{xiang2019generating} claimed that they are the first work to perform extensive adversarial attacks on 3D point cloud models by perturbing the positions of points or generating new points. Recent works extended the adversarial attacks for images to 3D point clouds by perturbing points and generating new points ~\citep{goodfellow2015explaining,yang2021adversarial,lee2020shapeadv}. Additionally, \citet{cao2019adversarial} proposed the adversarial attacks on LiDAR systems.\citet{wicker2019robustness} used occlusion attack and \citet{zhao2020isometry} proposed isometric transformations attack. Towards these adversarial attacks, corresponding defense techniques are developed ~\citep{yang2021adversarial,zhou2019dup}, which are more effective than adversarial training such as ~\citep{liu2019extending,zhang2019defense}. 
However, \citet{sun2021adversarial} examined existing defense works and pointed out that those defenses are still vulnerable to more powerful attacks. Thus, \citet{lorenz2021robustness} proposed the first verification algorithm for 3D point cloud models with provable robustness bounds. 

\textbf{Verification on Neural Networks:}
The robustness verification aims to find a guaranteed region that any input in this region leads to the same predicted label. For image classification, the region is bounded by a $l_p$-norm ball with a radius of $\epsilon$, and the aim is to maximize the $\epsilon$. For point clouds models, we reformulate the goal to find the maximum $\epsilon$ that any distortion applied on the position of the point within the region cannot alter the predicted label. Numerous works have attempted to find the exact value in the image domain $\epsilon$ ~\citep{katz2017reluplex,tjeng2018evaluating,bunel2018unified}. Yet these approaches are designed for small networks. Some works focus on computing a certified lower bound for the $\epsilon$. To handle the non-linearity operations in the neural network, the convex relaxation is proposed to approximate the bounds for the ReLU layer ~\citep{salman2020convex}. \citet{wong2018provable} and \citet{krishnamurthy2018dual} introduced a dual approach to form the relaxation. Several studies computed the bounds via linear-by-linear approximation ~\citep{wang2018efficient,weng2018fast,zhang2018efficient} or abstract domain interpretation ~\citep{gehr2018ai2,singh2019abstract}.

While for the 3D point cloud models, only \citet{lorenz2021robustness} proposed the 3Dcertify. However, their method is time expensive and can not handle the JANet in the full PointNet models. Thus, in this paper, we build a more efficient and general verification framework to obtain the certified bounds.   
\begin{figure}[t]
    \begin{center}
    \includegraphics[scale=0.3]{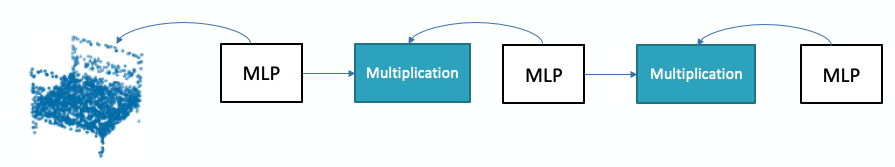}
    \end{center}
    \caption{Illustration of the combining forward and backward propagation process. The architecture in the MLP block contains convolution with ReLU activation function, batch normalization and pooling. Inside the MLP block, the bounds are computed by backward propagation.}
    \label{fig:combine}
    \vspace{-3mm}
\end{figure}
\section{Methodology: 3DVerifier}
\subsection{Overview}
The clean input point cloud $\mathbf{P}_0$ with $n$ points can be defined as {\footnotesize{$\mathbf{P}_0 = \{\mathbf{p_0}^{(i)} \mid\mathbf{p_0}^{(i)}\in \mathbb{R}^3, i = 1,..., n \}$}}, where each point $\mathbf{p_0}^{(i)}$ is represented in a 3D space coordinate $(x,y,z)$. We choose the points that are correctly recognized by the classifier $C$ as inputs to verify the robustness of $C$. 

Throughout this paper, we will perturb the points by shifting their positions in the 3D space within a distance bounded by the $l_p$-norm ball. Given the perturbed input {\footnotesize$\mathbf{P} = \{\mathbf{p}^{(i)} \mid\mathbf{p}^{(i)}\in \mathbb{R}^3, i = 1,..., n \}$} that is in the $l_p$-norm ball {\footnotesize{$\mathbb{S}_{p}\left(\mathbf{p}_{0}^{(i)}, \epsilon\right):=\left\{\mathbf{p}^{(i)} \mid\left\|\mathbf{p}^{(i)}-\mathbf{p}_{0}^{(i)}\right\|_{p} \leq \epsilon,  i = 1,..., n \right\}$}}, we aim to verify whether the predicted label of the model is stable within the region $\mathbb{S}_{p}$. This can be solved by finding the \textbf{\em minimum adversarial perturbation $\epsilon_{min}$} via binary search, such that {\footnotesize{$\exists \mathbf{P} \in \mathbb{S}_{p}\left(\mathbf{P}_0, \epsilon_{min}\right)$, argmax $C(\mathbf{P}) \neq c$}}, where $c=$argmax $C(\mathbf{P_0})$. Such $\epsilon_{min}$ is also referred as the {\em untargeted} robustness. As for the {\em targeted} robustness, it can be interpreted as that the prediction output score for the true class is always greater than that for the target class.

Assuming that the target class is $t$, the objective function is:
\begin{footnotesize}
\begin{center}
\begin{equation}
    \min\left\{ y_{c}(\mathbf{P})-y_{t}(\mathbf{P})\right\}:=\sigma_{\epsilon}, \qquad
    s.t \quad \| \mathbf{p}^{{k}}-\mathbf{p}_{0}^{{k}} \|_{p} \leq \epsilon, (k = 1,2,...,n),\label{con:eq1}
\end{equation}
\end{center}
\end{footnotesize}where $y_{c}$ represents the logit output for class $c$ and $y_{t}$ is for the target class $t$. $\mathbf{P}$ is the set of points that is centred around the original points set $\mathbf{P}_0$ within the $l_p$-norm ball with radius $\epsilon$. Thus, if $\sigma_{\epsilon} > 0$, the logit output of the true class $c$ is always greater than the target class, which means that the predicted label cannot be $t$. Due to the fact that finding the exact output of $\sigma_{\epsilon}$ is an NP-hard problem~\citep{katz2017reluplex}, the objective function of our work can be alternatively altered as computing the lower bound of $\sigma_{\epsilon}$. By applying binary search to update the perturbation $\epsilon$, we can find the minimum adversarial perturbation. Equivalently, the {\em maximum $\epsilon_{cert}$ }that does not alter the predicted label can be attained. Thus, in this paper, we aim to compute the \textbf{\em certified lower bound} of $\sigma_{\epsilon_{cert}}$.  
\subsection{Generic Framework}
To obtain the lower bound of $\sigma_{\epsilon}$ for the $l_p$-norm ball bounded model, we propagate the bound of each neuron layer by layer. As mentioned previously, most structures employed by the 3D point cloud models are similar to the traditional image classifiers, such as the convolution layer, batch normalization layer, and pooling layer. The most distinctive structure of the point clouds classifier, like the PointNet ~\citep{qi2017pointnet}, is the JANet structure. Thus, to compute the logit outputs of the neural network, our verification algorithm adopt three types of formulas to handle different operations: 1) linear operations (e.g. convolution, batch normalization and average pooling), 2) non-linear operation (e.g. ReLU activation function and max pooling), 3) multiplication operation. 

Let $\Phi^l(\mathbf{P})$ be the output of neurons in the layer $l$ with point clouds input $\mathbf{P}$. The input layer can be represented as $\Phi^0(\mathbf{P}) = \mathbf{P}$. Suppose that the total number of layers in the classifier is $m$, $\Phi^m(\mathbf{P})$ is defined as the output of the neural network. In order to verify the classifier, we aim to derive a linear function to obtain the global upper bound $u$ and lower bound $l$ for each layer output $\Phi^l(\mathbf{P})$ for $\mathbf{P} \in \mathbb{S}_{p}\left(\mathbf{P}_{0}, \epsilon\right) $.

The bounds are derived layer by layer from the first layer to the final layer. We have full access to all parameters of the classifier, such as weights $\mathbf{W}$ and bias $\mathbf{b}$. To calculate the output for each neuron, we show below the linear functions to obtain the bounds of the $l$-th layer for the neuron in position $(x,y)$ based on the previous $l'$-th layer: 
{\footnotesize{
\begin{equation}
\sum\limits_{i, j} \mathbf{A}_{(x, y, i, j)}^{\left(l, l^{\prime}\right), L} \Phi_{(x+i, j)}^{l^{\prime}}(\mathbf{P})+\mathbf{B}_{(x, y)}^{\left(l, l^{\prime}\right), L}  \leq \Phi_{(x, y)}^{l}(\mathbf{P}) \leq \sum\limits_{i, j} \mathbf{A}_{(x, y, i, j)}^{\left(l, l^{\prime}\right), U} \Phi_{(x+i, j)}^{l^{\prime}}(\mathbf{P})+\mathbf{B}_{(x, y)}^{\left(l, l^{\prime}\right), U}, \label{con:eq2}
\end{equation}}}where $\mathbf{A}^L, \mathbf{B}^L, \mathbf{A}^U, \mathbf{B}^U$ are weight and bias matrix parameters of the linear function for lower and upper bound calculations respectively. $\mathbf{A}$ and $\mathbf{B}$ are initially assigned as identity matrix ($\mathbf{I}$) and zeros matrix ($\mathbf{0}$), respectively, to keep the same output of $\Phi_{(x, y)}^l$ when $l=l'$. To calculate the bounds of current layer, we take backward propagation to previous layers. The $\Phi_{(x+i, j)}^{l^{\prime}}$ is substituted by the linear function of the previous layer recursively until it reaches the first layer ($l^\prime = 0$). After that, the output of each layer can be formed by a linear function of the first layer ($\Phi^0(\mathbf{P}) =\mathbf{P}$), as:{
{\footnotesize {\begin{equation}
\mathbf{A}^{(l,0),L} * \mathbf{P}+\mathbf{B}^{(l,0),L} \leq \Phi^{l}(\mathbf{P}) \leq \mathbf{A}^{(l,0),U} * \mathbf{P}+\mathbf{B}^{(l,0),U}. \label{con:eq3}
\end{equation}}}} Since the perturbation added to the point clouds input is bounded by the $l_p$-norm ball, $\mathbf{p} \in \mathbb{S}_{p}\left(\mathbf{p}_{0}, \epsilon\right)$, to compute the global bounds we need to minimize the lower bound and maximize the upper bound in Eq. \ref{con:eq3} over the input region. Thereby, the linear function to compute the global bounds for the $l$-th layer can be represented as:
{\footnotesize{\begin{equation}
\Phi_{x,y}^{l, U/L}=\pm \epsilon \|\mathbf{A}_{(x, y,:, :)}^{\left(l, 0\right), U/L}\|_{q}+\sum\limits_{i, j} \mathbf{A}_{(x, y, i, j)}^{\left(l, 0\right), U/L} \mathbf{p}_{0}^{(x+i, j)}+\mathbf{B}_{(x, y)}^{\left(l, 0\right), U/L} , \label{con:eq4}
\end{equation}
}}where $\|\mathbf{A}\|_q$ is the $l_q$-norm of $\mathbf{A}$ and $1/p+1/q =1$ with $p,q>1$, ``U/L" denotes that the equations are formulated for the upper bounds and lower bounds, respectively. This generic framework resembles to CROWN ~\citep{zhang2018efficient}, which is widely utilised in the verification works to verify feed-forward neural networks (e.g. CNN-Cert ~\citep{boopathy2018cnncert}, Transformer Verification ~\citep{Shi2020Robustness}). Unlike existing frameworks based on CROWN, we further extend the algorithm to verify point clouds classifiers.

\subsection{Functions for linear and non-linear operation}
As the linear and nonlinear functions are basic operations in the neural network, we first adapt the framework given in ~\citep{boopathy2018cnncert} to 3D point cloud models. In section 3.4, we will present our novel technique for JANet.

\textbf{Functions for linear operation.}
Suppose the output of the $l'$-th layer, $\Phi^{l'}(\mathbf{P})$, can be computed by the output of the ($l'$-1)-th layer, $\Phi^{l'-1}(\mathbf{P})$, by the linear function {\scriptsize{$\Phi^{l'}(\mathbf{P})=\mathbf{W}^{l'} * \Phi^{l'-1}(\mathbf{P}) + \mathbf{b}^{l'}$}}, where {\scriptsize{$\mathbf{W}^{l'}$}} and {\scriptsize{$\mathbf{b}^{l'}$}} are parameters of the function in the layer $l'$. Thus the Eq. \ref{con:eq2} can be propagated from the layer $l$ to the layer $l'-1$ by substituting $\Phi^{l^{\prime}}(\mathbf{P})$.

{\footnotesize{
$$
\setlength{\abovedisplayskip}{2pt}
\setlength{\belowdisplayskip}{2pt}
\begin{array}{l}
\Phi_{(x, y)}^{l,U/L}(\mathbf{P}) = \sum\limits_{i, j} \mathbf{A}_{(x, y, i, j)}^{\left(l, l^{\prime}\right), U/L} \Phi_{(x+i, j)}^{l^{\prime}}(\mathbf{P})+\mathbf{B}_{(x, y)}^{\left(l, l^{\prime}\right), U/L} \\
=\sum\limits_{i, j}\mathbf{A}_{(x, y, i, j)}^{\left(l, l^{\prime}\right), U/L}(\sum\limits_{k}\mathbf{W}^{l'}_{(k,j)} \Phi^{l'-1}_{(x+i,k)}(\mathbf{P}) + \mathbf{b}^{l'}_{x+i,j})+\mathbf{B}_{(x, y)}^{\left(l, l^{\prime}\right), U/L}\\
=\sum\limits_{i, k}\mathbf{A}_{(x, y, i, k)}^{\left(l, l^{\prime}-1\right), U/L}  \Phi^{l'-1}_{(x+i,k)}(\mathbf{P}) +\mathbf{B}_{(x, y)}^{\left(l, l^{\prime}-1\right), U/L}
\end{array}
$$
}}
\textbf{Functions for basic non-linear operation.} \label{sec:nonlinear}
For the $l'$-th layer with non-linear operations, we apply two linear functions to bound {\scriptsize{$\Phi^{\left(l^{\prime}\right)}(\mathbf{P})$}}:
{\footnotesize{
$$
\setlength{\abovedisplayskip}{2pt}
\setlength{\belowdisplayskip}{2pt}
\alpha^ {l^{\prime},L} \Phi^{l^{\prime}-1}(\mathbf{P})+\beta^{l^{\prime}, L} \leq \Phi^{\left(l^{\prime}\right)}(\mathbf{P}) \leq \alpha^{l^{\prime}, U} \Phi^{l^{\prime}-1}(\mathbf{P})+\beta^{l^{\prime}, U}.
$$
}}Given the bounds of {\scriptsize{$\Phi^{\left(l^{\prime}-1\right)}(\mathbf{P})$}}, the corresponding parameter {\footnotesize{$\alpha^L,\alpha^U,\beta^L,\beta^U$}} can be chosen appropriately. The functions to obtain the parameters are presented in Appendix A.

After computing the corresponding {\footnotesize{$\alpha^L,\alpha^U,\beta^L,\beta^U$}}, we back propagate the Eq.\ref{con:eq2} to the ($l'$-1)-th layer:
{\footnotesize{
$$
\begin{array}{l}
\setlength{\abovedisplayskip}{1pt}
\setlength{\belowdisplayskip}{1pt}
\Phi_{(x, y)}^{l,U/L}(\mathbf{P})= \sum\limits_{i, j} \mathbf{A}_{(x, y, i, j)}^{\left(l, l^{\prime}\right), U/L} \Phi_{(x+i, j)}^{l^{\prime}}(\mathbf{P})+\mathbf{B}_{(x, y)}^{\left(l, l^{\prime}\right), U/L} \\
= \sum\limits_{i, j} \mathbf{A}_{(x, y, i, j)}^{\left(l, l^{\prime}\right), U/L,+} *(\alpha^ {l^{\prime},U/L}_{(x+i,j)} \Phi^{l^{\prime}-1}(\mathbf{P})_{(x+i,j)}+\beta_{(x+i,j)}^{l^{\prime}, U/L})\\
+ \sum\limits_{i, j} \mathbf{A}_{(x, y, i, j)}^{\left(l, l^{\prime}\right), U/L,-} *(\alpha^ {l^{\prime},L/U}_{(x+i,j)} \Phi^{l^{\prime}-1}(\mathbf{P})_{(x+i,j)}+\beta_{(x+i,j)}^{l^{\prime}, L/U})\\
= \sum\limits_{i, j} \mathbf{A}_{(x, y, i, j)}^{\left(l, l^{\prime}-1\right), U/L} \Phi_{(x+i, j)}^{l^{\prime}}(\mathbf{P})+\mathbf{B}_{(x, y)}^{\left(l, l^{\prime}-1\right), U/L},
\end{array}
$$
}}where if {\footnotesize{$\mathbf{A}_{(x, y, i, j)}^{\left(l, l^{\prime}\right), U/L}$}} is a positive element of {\footnotesize{$\mathbf{A}_{(:, :, :, j)}^{\left(l, l^{\prime}\right), U/L}$}}, then {\footnotesize{$\mathbf{A}_{(x, y, i, j)}^{\left(l, l^{\prime}\right), U/L,+} = \mathbf{A}_{(x, y, i, j)}^{\left(l, l^{\prime}\right), U/L}$}} and {\footnotesize{$\mathbf{A}_{(x, y, i, j)}^{\left(l, l^{\prime}\right), U/L,-}= 0$}}; otherwise, {\footnotesize{$\mathbf{A}_{(x, y, i, j)}^{\left(l, l^{\prime}\right), U/L,-} = \mathbf{A}_{(x, y, i, j)}^{\left(l, l^{\prime}\right), U/L}$}} and {\footnotesize{$\mathbf{A}_{(x, y, i, j)}^{\left(l, l^{\prime}\right), U/L,+}= 0$}}.  
\subsection{Functions for Multiplication Layer}
The most critical structure in the PointNet model is the JANet, which contains the multiplication layers. For the multiplication, assume that it takes the output of previous layer ({\scriptsize{$\Phi^{l^{\prime}-1}(\mathbf{P})$}}) and ($l'$-$r$)-th layer ({\scriptsize{$\Phi^{l^{\prime}-r}(\mathbf{P})$, r $\in [1,l']$}}) as inputs, the output of {\scriptsize{$\Phi^{l^{\prime}}(\mathbf{P})$}} can be calculated by: {\footnotesize{
$\Phi_{(x, y)}^{l'}(\mathbf{P}) = \sum\limits_{k=1}^{d_k} \Phi_{(x, k)}^{l^{\prime}-r}(\mathbf{P})* \Phi^{l'-1}_{(k,y)}(\mathbf{P})$}}, where $d_k$ is the dimension of {\scriptsize{$\Phi_{(x,:)}^{l^{\prime}-r}(\mathbf{P})$ and $\Phi^{l'-1}_{(:,y)}(\mathbf{P})$}}. As the Transformer contains multiplication operation in self-attention layers, inspired by the Transformer verifier ~\citep{Shi2020Robustness}, we proposed our algorithm for point clouds models.  

In the JANet, before the multiplication layer, there is one reshape layer and one pooling layer. To simplify the computation, we choose the {\footnotesize{$\Phi^{l'-1}$}} as the output of pooling layer, by using $h=d_k*(k-1)+y$ to represent the transformation in the reshape layer. The equation to compute {\footnotesize{$\Phi^{l^{\prime}-1}(\mathbf{P})$}} can be rewritten as: \begin{footnotesize}
$\Phi_{(x, y)}^{l'}(\mathbf{P}) = \sum\limits_{k=1}^{d_k} \Phi_{(x, k)}^{l^{\prime}-r}(\mathbf{P})* \Phi^{l'-1}_{(0,h)}(\mathbf{P}),$\end{footnotesize}
where $h=d_k*(k-1)+y$.

To obtain the bounds of the multiplication layer, we utilize two linear functions of the input {\footnotesize{$\mathbf{P}$}} to bound {\footnotesize{$\Phi^{l^{\prime}}(\mathbf{P})$}}:
{\footnotesize{
\begin{equation}
\mathbf{\Lambda}^{(l',0),L}_{(x,y,i,:)}\mathbf{p}^{(x+i)}+\Theta^{(l',0),L}_{(x,y)} \leq\Phi_{(x, y)}^{l'}(\mathbf{P})\leq \mathbf{\Lambda}^{(l',0),U}_{(x,y,i,:)}\mathbf{p}^{(x+i)}+\Theta^{(l',0),U}_{(x,y)} \label{con:eq5}
\end{equation}
}}where $\lambda$ and $\Theta$ are new parameters of linear functions to bound the multiplication.
\begin{theorem}\label{thm1}
Let $l_r$ and $u_r$ be the lower and upper bounds of the $(l'$-$r)$-th layer output $(\Phi^{l^{\prime}-r}(\mathbf{P})$, r $\in [1,l'])$; $l_1$ and $u_1$ be the lower and upper bounds of the $(l'$-$1)$-th layer output $(\Phi^{l^{\prime}-1}(\mathbf{P}))$. Suppose $a^L = l_1, a^U = u_1, b^L=b^U=l_r, c^L = -l_r * l_1,$ and $c^U = -l_r * u_1$, Then, for any point clouds input $\mathbf{P} \in \mathbb{S}_{p}\left(\mathbf{P}_{0}, \epsilon\right) $ :
{\footnotesize{
$$
a^L*\Phi_{(x, k)}^{l^{\prime}-r}(\mathbf{P})+b^L*\Phi^{l'-1}(\mathbf{P})+c^L \leq
\Phi_{(x, k)}^{l^{\prime}-r}(\mathbf{P})* \Phi^{l'-1}_{(k,y)}(\mathbf{P}) \leq a^U*\Phi_{(x, k)}^{l^{\prime}-r}(\mathbf{P})+b^U*\Phi^{l'-1}_{(0,h)}(\mathbf{P})+c^U,
$$
}}
Proof: see detailed proof in Appendix B.
\end{theorem}
 The bounds and corresponding bounds matrix of {\small{$\Phi^{l^{\prime}-r}(\mathbf{P})$}} and {\small{$\Phi^{l^{\prime}-1}(\mathbf{P})$}} can be calculated via back propagation. Given Theorem \ref{thm1}, the functions can be formed to compute the $\mathbf{\Lambda}^{(l',0),U/L}$ and $\Theta^{(l',0),U/L}$ for $\Phi_{(x, y)}^{l'}(\mathbf{P})$ in Eq.\ref{con:eq5} as:
{\footnotesize{$$
\setlength{\abovedisplayskip}{2pt}
\setlength{\belowdisplayskip}{2pt}
\mathbf{\Lambda}^{(l',0),U/L}_{(x,y,i,:)} = \sum\limits_{k} ( a_{(0,h)}^{U/L,+}\mathbf{A}^{(l'-r,0),U/L}_{(x,k,i,:)}+a_{(0,h)}^{U/L,-}\mathbf{A}^{(l'-r,0),L/U}_{(x,k,i,:)})+$$
$$
\setlength{\abovedisplayskip}{2pt}
\setlength{\belowdisplayskip}{2pt}
\sum\limits_{k}(b_{(x,k)}^{U/L,+}\mathbf{A}^{(l'-1,0),U/L}_{(0,h,i,:)}+b_{(x,k)}^{U/L,-}\mathbf{A}^{(l'-1,0),L/U}_{(0,h,i,:)}), $$
$$\Theta^{(l',0),U/L}_{(x,y)} = \sum\limits_{k} ( a_{(0,h)}^{U/L,+}\mathbf{B}^{(l'-r,0),U/L}_{(x,k)}+a_{(0,h)}^{U/L,-}\mathbf{B}^{(l'-r,0),L/U}_{(x,k)})+$$
$$
\setlength{\abovedisplayskip}{2pt}
\setlength{\belowdisplayskip}{2pt}
\sum\limits_{k}(b_{(x,k)}^{U/L,+}\mathbf{B}^{(l'-1,0),U/L}_{(k,y)}+b_{(x,k)}^{U/L,-}\mathbf{B}^{(l'-1,0),U/L}_{(0,h)})+c^{U/L}_{(x,y)}.$$
}}Thereby, it is a forward propagation process by employing the computed bounds metrics of $\Phi^{\left(l^{\prime}-r\right)}(\mathbf{P})$ and $\Phi^{\left(l^{\prime}-1\right)}(\mathbf{P})$ to obtain the bounds of $\Phi^{\left(l^{\prime}\right)}(\mathbf{P})$. When it comes to the later layer, the $l$-th layer, we use the backward process to propagate the bounds to the multiplication layer, which can be referred as the $l'$-th layer. Next, at the multiplication layer, we propagate the bounds to the input layer directly by skipping previous layers. The bounds that propagate to the multiplication layer ($l'$-th layer) can be represented by the Eq. \ref{con:eq2}. Therefore, $\Phi^{l'}(\mathbf{P})$ can be substituted by the linear functions in Eq. \ref{con:eq5} to obtain $\mathbf{A}^{(l,0),U/L}$ and $\mathbf{B}^{(l,0),U/L}$:
{\footnotesize{$$\mathbf{A}^{(l,0),U/L}_{(x,y,i,:)}=\sum\limits \left( \mathbf{A}^{(l,l'),U/L,+}_{(x,y,i,:)} \mathbf{\Lambda}^{(l',0),U/L}_{(x+i,y,:,:)}+\mathbf{A}^{(l,l'),U/L,-}_{(x,y,i,:)} \mathbf{\Lambda}^{(l',0),L/U}_{(x+i,y,:,:)}\right),$$ }}
{\footnotesize{$$ \mathbf{B}^{(l,0),U/L}_{(x,y)}= \mathbf{B}^{(l,l'),U/L}_{(x,y)} + \sum\limits (\mathbf{A}^{(l,l'),U/L,+}_{(x,y,i,:)} \mathbf{\Theta}^{(l',0),U/L}_{(x+i,y)} +\mathbf{A}^{(l,l'),U/L,-}_{(x,y,i,:)} \mathbf{\Theta}^{(l',0),L/U}_{(x+i,y)}).
 $$}}Lastly, the global bounds of the $l$-th layer can be computed using the linear functions in Eq. \ref{con:eq4}.
 \begin{figure*}
    \begin{center}
    \includegraphics[width=0.8\textwidth]{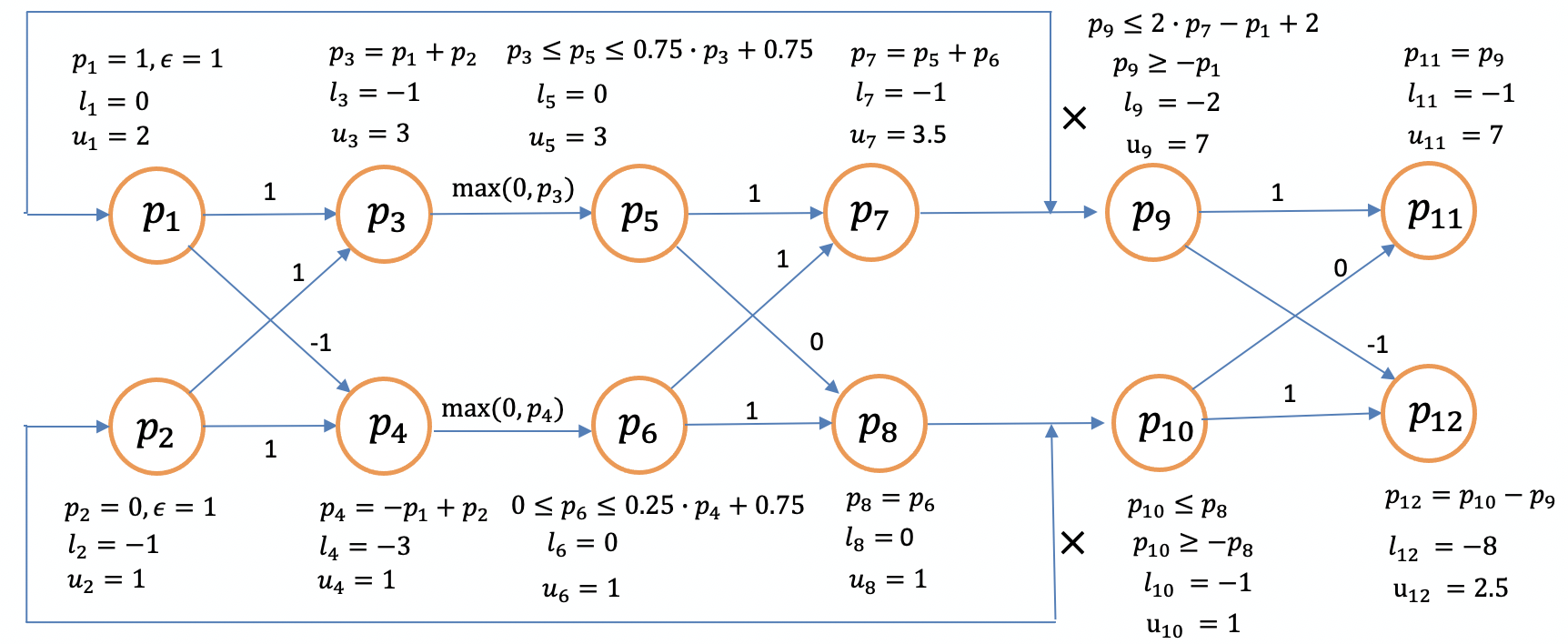}
    \end{center}
    \caption{A running example for a simple neural network with multiplication. The inputs ($p_1$, $p_2$) are bounded by a $l_\infty$-norm ball with radius $\epsilon$.}
    \label{fig:example}
    \vspace{-3mm}
\end{figure*}
\subsection{A Running Numerical Example}
To demonstrate the verification algorithm, we present a simple example network in Fig. \ref{fig:example} with two input points, $p_1$ and $p_2$. Suppose the input points are bounded by a $l_\infty$-norm ball with radius $\epsilon$, our goal is to compute the lower and upper bounds of the outputs ($p_{11},p_{12}$) based on the input intervals. As Fig. \ref{fig:example} shows, the neural network contains 12 nodes, and each node is assigned a weight variable. There are three types of operations in the example network: linear operation, non-linear operation (ReLU activation function) and multiplication. 

Given the input points $\mathbf{P} = [p_1=1,p_2=0]$, to obtain the bounds for $p_3$ and $p_4$, according to the Eq. \ref{con:eq2}, $\mathbf{A}^{(1,0)}$ can be assigned as
$\mathbf{A}^{(1,0)}_{(0,0,:,:)} = $ 
$\begin{bmatrix}
    1       & 1  \\
\end{bmatrix}$ 
and $\mathbf{A}^{(1,0)}_{(0,1,:,:)} = $ 
$\begin{bmatrix}
    -1       & 1  \\
\end{bmatrix}$
to compute the output for the 1-st layer $[p_3,p_4]$: {\footnotesize
\begin{equation}
    \begin{array}{cc}
         p_3 = p_1 \cdot \mathbf{A}^{(1,0)}_{(0,0,0,0)}+p_2\cdot\mathbf{A}^{(1,0)}_{(0,0,0,1)} = 1 , 
         p_4 = p_1 \cdot \mathbf{A}^{(1,0)}_{(0,1,0,0)}+p_2\cdot\mathbf{A}^{(1,0)}_{(0,1,0,1)} = -1.
    \end{array}\label{con:eq6}
\end{equation}}

Based on Eq.\ref{con:eq4}, we can obtain the lower bound and upper bound for the perturbed input ($p =\infty$): $l_3 = 1-2\epsilon = -1$, $u_3 = 1+2\epsilon = 3$, $l_4 = -1-2\epsilon = -3$, and $u_4 = -1+2\epsilon = 1$.

For the non-linear ReLU activation layer, 
according to the relaxation rule demonstrated in Appendix A, in our example, for $p_5$, we obtain $\alpha_L = 1$, $\beta_L = 0$, $\alpha_U = 0.75$, and $\beta_U = 0.75$; for $p_6$, we choose  $\alpha_L = \beta_L = 0$, $\alpha_U = 0.25$, and $\beta_U = 0.75$. Then we can obtain the bounds for $p_5$ and $p_6$ as shown in Figure \ref{fig:example}.

 Next, for the 3-rd layer output $[p_7,p_8]$: $p_7 = p_5 + p_6$, $p_8 = p_6$, we assign $\mathbf{A}^{(3,2),L}_{(0,0,:,:)} = \mathbf{A}^{(3,2),U}_{(0,0,:,:)} = [1,1]$ and $\mathbf{A}^{(3,2),L}_{(0,1,:,:)} = \mathbf{A}^{(3,2),U}_{(0,1,:,:)} = [0,1]$ to form:
 {\footnotesize
 $$ p_5 \cdot \mathbf{A}^{(3,2),L}_{(0,0,0,0)} + p_6 \cdot \mathbf{A}^{(3,2),L}_{(0,0,0,1)} \leq p_7 \leq p_5 \cdot \mathbf{A}^{(3,2),U}_{(0,0,0,0)} + p_6 \cdot \mathbf{A}^{(3,2),U}_{(0,0,0,1)}$$ 
 $$ p_5 \cdot \mathbf{A}^{(3,2),L}_{(0,1,0,0)} + p_6 \cdot \mathbf{A}^{(3,2),L}_{(0,1,0,1)} \leq p_8 \leq p_5 \cdot \mathbf{A}^{(3,2),U}_{(0,1,0,0)} + p_6 \cdot \mathbf{A}^{(3,2),U}_{(0,1,0,1)}.$$}
 We can back propagate constrains to the first layer to gain final global bounds of $p_7$ and $p_8$: 
 {\scriptsize
\begin{equation}
\begin{array}{c}p_{1}+p_{2} \leq p_{7} \leq 0.5 \cdot p_{1}+p_{2}+1.5 \\ 0 \leq p_{8} \leq-0.25 \cdot p_{1}+0.25 \cdot p_{2}+0.75.\end{array}\label{con:eq7}
\end{equation}}
Similarly, we compute the global bounds for $p_7$ and $p_8$.

For the multiplication layer, according to the formulations presented in section 3.4, to compute the bounds of $p_9$ in our example, we set $a^L = l_1$, $a^U = u_1$, $b^L = b^U = l_7$, $c^L = -l_7 \cdot l_1$, and $c^U = -l_7 \cdot u_1$. Similarly, for $p_9$, we choose $a^L = l_2$, $a^U = u_2$, $b^L = b^U = l_8$, $c^L = -l_8 \cdot l_2$, and $c^U = -l_8 \cdot u_2 $. Thus, we get
\begin{equation}
    \begin{array}{c}
         - p_1 \leq p_9 \leq 2 \cdot p_7 - p_1 + 2  ,\quad
         -p_8 \leq p_{10} \leq p_8.
    \end{array}\label{con:eq8}
\end{equation}
Instead of performing back-propagation to the input layer, we calculate the bounds for $p_9$ and $p_{10}$ by directly replacing $p_7$ and $p_8$ with Eq.\ref{con:eq7}: 
\begin{equation}
    \begin{array}{c}
         -p_1 \leq p_9 \leq  2 \cdot p_2 + 5  \\
          0.25 \cdot p_1 -0.25 \cdot p_2 -0.75 \leq p_{10} \leq -0.25 \cdot p_1 + 0.25 \cdot p_2 + 0.75. 
    \end{array}
\end{equation}
Again, according to Eq.\ref{con:eq4}, we obtain $l_9 = -2$, $u_9 = 7$, $l_{10} = -1$ and $u_{10} = 1$.

Finally, in our example, $p_{11} = p_9$ and $p_{12} = p_{10}-p_9$. After propagating the linear bounds to the previous layer by replacing the $p_9$ and $p_{10}$ with $p_7$ and $p_8$ in Eq. \ref{con:eq8}, we construct the constraints of the last layer as:
\begin{equation}
    \begin{array}{c}
          -p_1 \leq p_{11} \leq  2 \cdot p_7 - p_1 + 2   \\
         -p_8-2\cdot p_7 +p_1 -2 \leq p_{12} \leq  p_8+p_1.
    \end{array}
\end{equation}
By substituting $p_7$ and $p_8$ with $p_1$ and $p_2$ directly, the global bounds for $p_{11}$ and $p_{12}$ are:
\begin{equation}
    \begin{array}{c}
         -p_1 \leq p_{11} \leq  2 \cdot p_2 + 5   \\
         0.25 \cdot p_1 -2.25 \cdot p_2 -5.75 \leq p_{12} \leq  0.75 \cdot p_1 + 0.25 \cdot p_2 + 0.75.
    \end{array}
\end{equation}
Thereby, as described so far, the back-substitution yields $l_{11} = -1$, $u_{11} = 7$, $l_{12} = -8$, $u_{12} = 2.5$, which are the final output bounds of our example network. 

\textit{Robustness Analysis:} The inputs $p_1 = 1$ and $p_2 = 0$ lead to output $p_{11}= 1$ and $p_{12} = -1$ in our example. Thus, to verify the robustness of our example network, we aim to find the maximum $\epsilon$ that guarantees $p_{11} \geq p_{12}$ always holds true for any perturbed inputs within the $l_\infty$-norm ball with a radius $\epsilon$. In our example, the results of $l_{11}$, $u_{11}$, $l_{12}$, and $u_{12}$ conclude that $p_{11} - p_{12} \in [-3.5,3]$ which results in $p_{11} \geq p_{12}$ failing to hold. Thus, we apply binary search to reduce the value of $\epsilon$ and recalculate the output bounds for the network based on the new $\epsilon$. 
When the maximum iteration is reached, we stop the binary search and choose the maximum $\epsilon$ that results in the lower bound of $p_{11} - p_{12} \geq 0$ as the final certified distortion. 
\section{Experiments} 
\textbf{Dataset} 
We evaluate 3DVerifier on ModelNet40 ~\citep{wu20153d} dataset, which contains 9,843 training and 2,468 testing 3D CAD models. Each CAD model is used to sample a point cloud that comprises 2,048 points in three dimensions ~\citep{qi2017pointnet}. There are 40 categories of CAD models. We run verification experiments on point clouds with 1024, 512, and 64 points, and randomly selected 100 samples from all the 40 categories of the test set as the dataset to perform the experiments. All experiments are carried out on the same randomly selected dataset.

\textbf{Models} We utilize PointNet ~\citep{qi2017pointnet} models as the 3D point cloud classifiers. Since the baseline verification method, 3DCertify ~\citep{lorenz2021robustness}, cannot handle the full PointNet with JANet, to make a comprehensive comparison, we first perform experiments on the PointNet without JANet. We then examine the performance of our 3DVerifier on full PointNet models. All models use the ReLU activation function.   

\textbf{Baseline} (1) We choose the existing 3D robustness certifier, 3DCertify~\citep{lorenz2021robustness}, as the main baseline for the PointNet models without JANet, which can be viewed as general CNN. Additionally, we also show the average distortions obtained by adversarial attacks on 3D point cloud models that extended from the CW attack ~\citep{carlini2017evaluating,xiang2019generating} and PGD attack~\citep{kurakin2016adversarial}. (2) As for the complete PointNet proposed by~\citep{qi2017pointnet}, we provide the average and minimum distortions obtained by the CW attack for robustness estimation. The PGD attack takes a long time to seek the adversarial examples and the attack success rate is below 10\%, thus, it is not included as the comparative method. 

\textbf{Implementation}
The 3DVerifier is implemented via NumPy with Numba in Python. All experiments are run on a 2.10GHz Intel Xeon Platinum 8160 CPU with 512 GB memory.

\begin{table}[h]\footnotesize
\begin{center}
\caption{Average certified bounds (ave) and run-time on PointNet without JANet. For the certified bounds, the higher the better. The bounds obtained by attacks are referred as upper bounds. `*' means that computing the bounds by 3DCertify is computationally unattainable, which automatically terminates after verifying several samples. Thus, we present the results of verifying 10 samples in Appendix C.}
\label{label:results1}
\scalebox{0.8}{
\begin{tabular}{c|c|c|c|c|cc|cc|cc|c|c}

no. points&Pooling &Acc.& N  & $l_p$& \multicolumn{4}{c}{Certified Bounds} &\multicolumn{2}{|c|}{Our vs.}  & \multicolumn{2}{c}{Attack}  
\\ \cline{6-9} \cline{12-13} 
 & & & &  & \multicolumn{2}{c}{Our} & \multicolumn{2}{c|}{3DCertify} &\multicolumn{2}{c|}{3DCertify (\%)} &CW  & PGD   \\
 \cline{6-13} 
 & & & & &ave & time(s) & ave  & time(s) & ave  & time(s) & ave  & ave
 \\ \hline
64&Average&70.1& 5 & $l_\infty$ & 0.0122 & 0.51 & 0.0080 & 1.59&+52.5 &-67.9  & 0.0270&0.0222   \\
   & & & & $l_2$   & 0.0433 & 0.46&-& -&  &  &0.4207 & 0.5278 \\ 
& & & &{$l_1$}  & 0.0553 & {0.39} & -  &{-} &  & & {2.5030}   & {3.2103}   
\\ \cline{3-13}
& & 79.25 & 9& $l_\infty$ & 0.0044 & 8.87 & 0.0034& 92.08&+29.4 &-90.4 &0.0231 & 0.0166\\
& & & &$l_2$   & 0.0146 & 6.84  & -   & - &  &  & 0.3873  & 0.4624\\                  
& & & &$l_1$  & 0.0188& 6.71  &-  & -&  & & 2.9722 &2.5512         
\\ \cline{2-13}
 &Max &73.55 & 5& $l_\infty$ & 0.0137 & 0.40 & 0.0079 & 0.544&+73.4 &-26.5 & 0.0342& 0.0222 
\\
 &  & & &$l_2$   & 0.0481 & 1.53&-& -& & &0.1558 & 0.2786              
\\ 
& & & &{$l_1$}  & 0.0649 & {0.94} &-  &{-} & & &  {3.1079}   & 3.2104    
\\ \cline{3-13}
 &&81.22&9&$l_\infty$ & 0.0051 & 8.33 & 0.0036  & 51.43&+41.7 &-83.8 &0.0201 &0.0277  
\\
 &  & & &$l_2$   & 0.0209 & 10.48&-& -& & &0.2366 & 0.1587              
\\ 
& & & &{$l_1$}  & 0.0288 & {9.45} &-  &{-} & & &  {2.8055}   &  3.0901   
 \\ \hline
 512&Average  &73.02&5& $l_\infty$ & 0.0104 & 15.27 & 0.0063 & 100.44&+15.5 &-84.8 & 0.0705&0.0221   
\\
   & & & &$l_2$   & 0.0427 & 13.21&-& -& & &0.4051 & 0.4733              
\\ 
& & & &{$l_1$}  & 0.0648 & {12.32} &-  &{-} & & & {4.5381}   &   4.4081
\\ \cline{2-13}
 & Max  &78.32&5& $l_\infty$& 0.0108 & 7.65 &0.0070 & 18.52&+54.3 & -58.7 &0.0572  & 0.0243 
\\
   & && &$l_2$   &0.0313  &5.19 &-& -&- & -& 0.2576 & 0.1870              
\\ 
& & && {$l_1$}  & 0.0407 & {3.76} &-  &- & & & {2.8966}   & 3.1079  
\\ \cline{3-13}
& &82.12&9& $l_\infty$ & 0.0048 &  60.22& * &* & -& -& 0.0296 & 0.0153 
\\
   & & & &$l_2$   & 0.0159  &58.78 &-& -&- & -&0.2627 & 0.3583              
\\ 
& & & &{$l_1$}  & 0.0196 &  {54.34} &-  &{-} &  & & {4.4918}   & 5.0793  
\\ \hline
1024&Average  & 72.61&5& $l_\infty$ & 0.0145 & 43.05 &* &* &- &- &0.0475 & 0.0356 
\\
   & & &&$l_2$   &0.0471  &21.88 &-& -&- & -& 0.4635& 0.3877              
\\ 
& & & &{$l_1$}  & 0.0544 &  {10.08} &-  &{-} & & & {2.3798}   & 3.0146  
\\ \cline{2-13}
&Max  &77.83 &5& $l_\infty$ & 0.0135  & 14.37 &0.0085  & 21.01 &+58.8 &-31.6 &0.0642 & 0.0345 
\\
   & & &&$l_2$   &0.0393  &19.01 & -& - & -& -&0.3629 & 0.2764           
\\ 
& & & &{$l_1$}  &0.0500  & {18.07} &-  &{-} &- &- & {3.4692}   & 2.4467
\\ \hline
\end{tabular}}
\end{center}
\end{table}
\subsection{Results for PointNet models without JANet}
In Table \ref{label:results1}, we present the clean test accuracy (Acc.) and average certified bounds (ave) for PointNet models without JANet. We also record the time to run one iteration of the binary search. We demonstrate that our 3DVerifier can improve upon 3DCertify in terms of run-time and tightness of bounds. To make the comparison more extensive, we train a 5-layer ($N$=10) model and a 9-layer ($N$=25) model, which include two multilayer perceptrons (MLPs) and three MLPs with one pooling layer, respectively. We also show the performance of models with different types of pooling layers: global average pooling and global maximum pooling. For experiments, we set the initial $\epsilon$ as 0.05 and the maximum iteration of binary search as 10. The experiments are taken on three point-cloud datasets with 64, 512, and 1024 points. We can see from the results of average bounds that our 3DVerifier can obtain tighter lower bounds than 3DCertify and engages a significant gap on the distortion bounds found by CW and PGD attacks. Table \ref{label:results1} shows that our method is much faster than the 3DCertify. Notably, 3DVerifier enables an orders-of-magnitude improvement in efficiency for large networks. Additionally, our method can compute certified bounds of distortion constrained by $l_1$, $l_2$, and $l_\infty$-norms, which is more scalable than 3DCertify in terms of norm distance.
\begin{table}\footnotesize
\begin{center}
\caption{Average certified bounds (ave) and times on PointNet models with T-Net}
\label{label:results}
\scalebox{1}{\begin{tabular}{c|c|c|c|c|cc|c|cc}
\hline
no.points &Pooling& Acc. & N &{$l_p$}& \multicolumn{3}{c|}{Our}   &\multicolumn{2}{c}{CW}     
\\ \cline{6-10} 
 & & & & & ave&min& time (s) & ave & min
 \\ \hline
64 & Average & 70.71& 5&$l_\infty$  & 0.341 & {0.021} & 0.12  &0.727 & {0.011}
\\ 
&  & & &$l_2$   &0.862 &0.077& 0.19& 1.700 & {0.162}  
\\ 
 & & & &{$l_1$} &   1.479 &{0.136} & 0.18& 2.049  & {0.317}
\\ \cline{3-10}
& & 80.22 &9&$l_\infty$  &0.316 &{0.016}&2.98 & 0.832 &{0.053}
\\ 
 & & && {$l_2$}   &  0.588 &{0.032}&  2.86  &2.164 & {0.276} 
\\ 
{} & & &{}& {$l_1$} & 1.115 &{0.113} & 2.73 & 2.949 & {0.510} 
\\ \cline{3-10}
& & {84.11} &13&$l_\infty$  & 0.248 & {0.012}&23.98  &0.792  & {0.077}
\\ 
{} &  &{}& & {$l_2$}   & 0.466 &{0.026}& 23.25 & 3.310 & {0.313} 
\\ 
{} & & & {}&{$l_1$} &  0.857 &{0.088} & 22.79&  4.857   &{0.875}
\\ \cline{2-10}
& Max & {74.63} &5&$l_\infty$ &0.554 & {0.029}&0.43 &0.791 & {0.093}
\\ 
{} &  &{}& & {$l_2$} &1.118  &{0.039} &0.41 & 2.133 & {0.137} 
\\ 
{} & & {}& &{$l_1$} & 1.650  &{0.041} &0.397 &    3.078 &0.322 
\\ \cline{3-10}
&  &81.91 &9&$l_\infty$  & 0.541 & {0.007}&4.26 &0.846  & {0.078}
\\ 
{} & & &{}& {$l_2$}   & 0.897 & {0.010} &4.11 & 2.554 & {0.102} 
\\ 
{} & & &{}&{$l_1$} & 1.326  &{0.010} &3.18 & 3.772    &{0.167}
\\ \cline{3-10}
 & &86.77 &{13}&$l_\infty$  &0.122  & {0.027}&24.42 &0.218 & 0.044
\\ 
{} & & &{} &{$l_2$}   &0.494  &{0.030}&24.13 &1.285  & {0.076} 
\\ 
{} & & &{}&{$l_1$} & 0.615  &{0.035} &23.29 &1.721 &{0.103} 
\\ \hline
512 & Average & 73.53&{9}&$l_\infty$  &{0.231}&0.005 & 71.26  &0.395  & {0.035}
\\ 
{} & & &{}& {$l_2$}   &1.196  &{0.027}&68.73 & 3.106 & {0.637}  
\\ 
{} & & & {}&{$l_1$} & 1.622  &{0.107} &66.91  &4.852    & {0.833} 
\\ \cline{2-10}
 & Max & 84.61&{9}&$l_\infty$  &0.109  & {0.0003}&55.78 &0.693  & {0.033}
\\ 
{} & & &{}& {$l_2$}   & 0.272 &{0.0005}&53.64 & 0.882 & {0.085} 
\\ 
{} & && {}&{$l_1$} & 0.345  &{0.043} & 50.22 & 1.175 & {0.448} 
\\ \hline
1024 & Average & 71.47&9&$l_\infty$  &0.406  & {0.017}&129.41 & 0.721 & {0.041}
\\ 
{} & & &{}& {$l_2$}   & 1.142 &{0.181}& 127.88& 1.459 & {0.342}  
\\ 
{} & & &{}&{$l_1$} &1.574 &{0.233} &126.24 & 1.926&{0.596}
\\ \cline{2-10}
 & Max&80.22 &9&$l_\infty$  &0.374  & {0.001}& 128.22& 0.758 & {0.005}
\\ 
{} & & &{}& {$l_2$}   & 0.959 &{0.006}&127.51 & 1.591 & {0.009} 
\\ 
{} & & & {}&{$l_1$} & 1.253  &{0.009} &120.75 & 2.163 &{0.018}
\\ \hline
\end{tabular}}
\end{center}
\end{table}

\subsection{Results for PointNet models with JANet}
As the 3DCertify does not include the bounds computation algorithm for multiplication operation, it can not be applied to verify the complete PointNet with JANet architecture. Thus, as the first work to verify the point cloud models with JANet, we show the average and minimum distortion bounds of CW attack-based method to make comparisons. We examine $N$-layer models ($N$= 5,9,13) with two types of global pooling: average pooling and maximum pooling on datasets with 64, 512, and 1024 points. The obtained certified average bounds of full PointNet models are shown in Table \ref{label:results}. According to previous verification works on images (e.g. ~\citep{boopathy2018cnncert}) and results in Table \ref{label:results1}, the gap between certified bounds and attack-based average distortion is reasonable, where the average minimum distortion are 10 times greater than the bounds. Thus, it reveals that our method efficiently certifies the point cloud models with JANet in a comparable quality with models without JANet.
\subsection{{Discussions}}
\noindent\textbf{3DVerifier is efficient for large-scale 3D point cloud models.}\\
There are two key features that enable 3DVerifier's efficiency for large 3D networks.

\textit{Improved global max pooling relaxation:} 
the relaxation algorithm for global max pooling layer of CNN-Cert ~\citep{boopathy2018cnncert} framework cannot be adapted to 3D point cloud models directly, which is computationally unattainable. Thus, we proposed a linear relaxation for the global max pooling layer based on ~\citep{singh2019abstract}. For example, to find the maximum value $p_r$ = $\mathop{max}_{m \in \mathcal{M}}(p_m)$, if exists $p_j$ such that its lower bound $l_j \geq u_k$ for all $k \in \mathcal{M} \backslash {j}$, the lower bound for the max pooling layer is $l^r=l_j$ and upper bound is $u^r =u_j$. Otherwise, we set the output of the layer $\phi^r \geq p_j $, where $j = \mathop{argmax}\limits_{m \in \mathcal{M}}(l_m)$, and similarly $\phi^r \leq p_k $, where $k = \mathop{argmax}\limits_{m \in \mathcal{M}}(u_m)$. The comparison results in Table \ref{label:results1} indicates that implementing the improved relaxation for the max pooling layer enables 3DVerifier to compute the certified bounds much faster than the existing method, 3DCertify.

\textit{Combing forward and backward propagation:} The verification algorithm proposed in ~\citep{Shi2020Robustness} evaluated the effectiveness of combining forward and backward propagation to compute the bounds for Transformer with self-attention layers. Thus, in our tool 3DVerifier, we adapted the combining forward and backward propagation to compute the bounds for the multiplication layer. As Table \ref{label:results} shows, the time spent for full PointNet models with JANet is nearly the same as the time for models without JANet.

\textbf{CNN-Cert can be viewed as a special case of 3DVerifier.}\\
CNN-Cert ~\citep{boopathy2018cnncert} is a general efficient framework to verify the neural networks for image classification, which employs 2D convolution layers. Although our verification method shares a similar design mechanism as CNN-Cert, our framework is superior to CNN-Cert. One key difference is the dimension of input data, PointNet 3D models adopt 1D convolution layers, which can be efficiently handled by 3DVerifier. Additionally, besides the general operations such as pooling, batch normalization, and convolution with ReLU activation, we can also handle models with JANet that contains multiplication layers. Thus, 3DVerifier can tackle more complex and larger neural networks than CNN-Cert. To adapt the framework for 3D point clouds, we introduced a novel relaxation method for max-pooling layers to obtain its certified bounds, leading to a significant improvement in terms of efficiency.

\begin{table}
\begin{center}
\caption{Training accuracy of different structures of point cloud models.}
\label{label:accJAnet}
\begin{tabular}{|l|c|c|c|}
\hline
Models & N&no. trainable parameters & accuracy  \\
\hline\hline
PointNet without JANet &7& 443656 & 72.94\% \\
\hline
PointNet without JANet & 13 &822472 &74.07\% \\
\hline
Full PointNet with JANet & 13 & 645425 & 81.31\%\\
\hline
\end{tabular}
\end{center}
\end{table}
\section{The Power of JANet}
We perform ablation study to show the importance of JANet in the point clouds classification task. We choose three models to train, and record the number of trainable parameters for each model. The final results of training accuracy are demonstrated in Table \ref{label:accJAnet}. The specific layer configurations for these models are presented in Appendix D. As the PointNet without JANet can be regarded as a general CNN model, while JANet is a complex architecture that contains T-Net and multiplication layer, to interpret the effect of JANet, we examine two models for the PointNet without JANet: 1) 7-layer model, which abandons the JANet in the 13-layer full PointNet; 2) 13-layer model, which intuitively adds the convolution and dense layers in T-Net to the 7-layer model. From the results, we can see that the PointNet with JANet can improve the training accuracy significantly comparing with PointNet without JANet when they employ the same number of layers. Therefore, JANet plays an important role to improve the performance of point clouds models.

\section{Conclusion}
In this paper, we proposed an efficient and general robustness verification framework to certify the robustness of 3D point cloud models. By employing the linear relaxation for the multiplication layer and combining forward and backward propagation, we can efficiently compute the certified bounds for various model architectures such as convolution, global pooling, batch normalization, and multiplication. Our experimental results on different models and point clouds with a different number of points confirm the superiority of 3DVerifier in terms of both computational efficiency and tightness of the certified bounds. 

\begin{appendices}
\section{Relaxations for Non-linearity}
In section 3.3, we mentioned that to compute bounds for non-linear functions, we choose appropriate parameters to form the linear relaxation to bound the layer containing nonlinear operations. Suppose that the non-linear function is $f(y)$, where $y$ is the output from previous layer $\Phi(\mathbf{P})^{l^{\prime}-1}$. The bound obtained from the previous layer is $[l,u]$. Thus, our goal is to bound the $f(y)$ by the following constraints:
$\alpha^ {l^{\prime},L} y +\beta^{l^{\prime}, L} \leq f(y) \leq \alpha^{l^{\prime}, U} y+\beta^{l^{\prime}, U},$
where parameters $\alpha^ {l^{\prime},L}, \beta^{l^{\prime}, L}, \alpha^{l^{\prime}, U}$, and $\beta^{l^{\prime}, U}$ are chosen dependent on the lower and upper bound, $l$ and $u$, of the previous layer, which follow different rules for different functions.

The way to bound the non-linear functions has been thoroughly discussed in previous works on images (e.g. ~\citep{zhang2018efficient,boopathy2018cnncert,singh2019abstract}). Thus, by reviewing their methods, we synthesise our relaxations for the nonlinear functions to determine the parameters according to $l$ and $u$.

\textbf{ReLU} The most common activation function is the ReLU function, which is represented as $f(y) = max(0,y)$. Thus, if $u \leq 0$, the output of $f(y)$ is 0; if $l \geq 0$, the output will exactly be $y$. As for the situation that $l<0$ and $u>0$, we set the upper bound $u$ to be the line cross two endpoints: $(l,0)$ and $(u, f(y))$, which can be represented as 
$f^U(y) = \frac{u(y-l)}{(u-l)}$
As for the expression for the lower bound, to make the bounds tighter, we consider two cases: if $u>\lvert l\rvert$, $f^L(y) = y$ and otherwise $f^L(y) = y$.
Therefore, we can choose the $\alpha^U = \frac{u}{(u-l)}, \beta^U = \frac{-ul}{(u-l)} , \beta^L = 0$; $\alpha^L = 0$ when $u<\lvert l\rvert$, and otherwise $\alpha^L = 1$. 
\section{Derivation of the linear functions to bound the multiplication layer}
The mathematical proof to choose the optimal parameters of the linear functions to bound the multiplication layer is evaluated in \cite{Shi2020Robustness}. Here we will present how to derive the optimisation problem based on the derivation in \cite{Shi2020Robustness}. The goal of the linear relaxation is to bound the multiplication on two inputs, $p_1p_2$, which can be represented by the following constraints:
\begin{equation}
    a^L p_1 + b^L p_2 + c^L \leq p_1p_2 \leq a^Up_1 + b^Up_2 + c^U
\end{equation}
As we have already obtained the bounds of $p_1$ and $p_2$, given the bounds of $p_1$ is $[l_1, u_1]$ and for $p_2$ is $[l_2, u_2]$, we can choose the appropriate parameters of $a^L$, $a^U$, $b^L$, $b^U$, $c^L$ and $c^U$ according to the computed bounds.

\subsection{derivation for the lower bound}
\begin{figure}
    \begin{center}
    \includegraphics[scale=0.1]{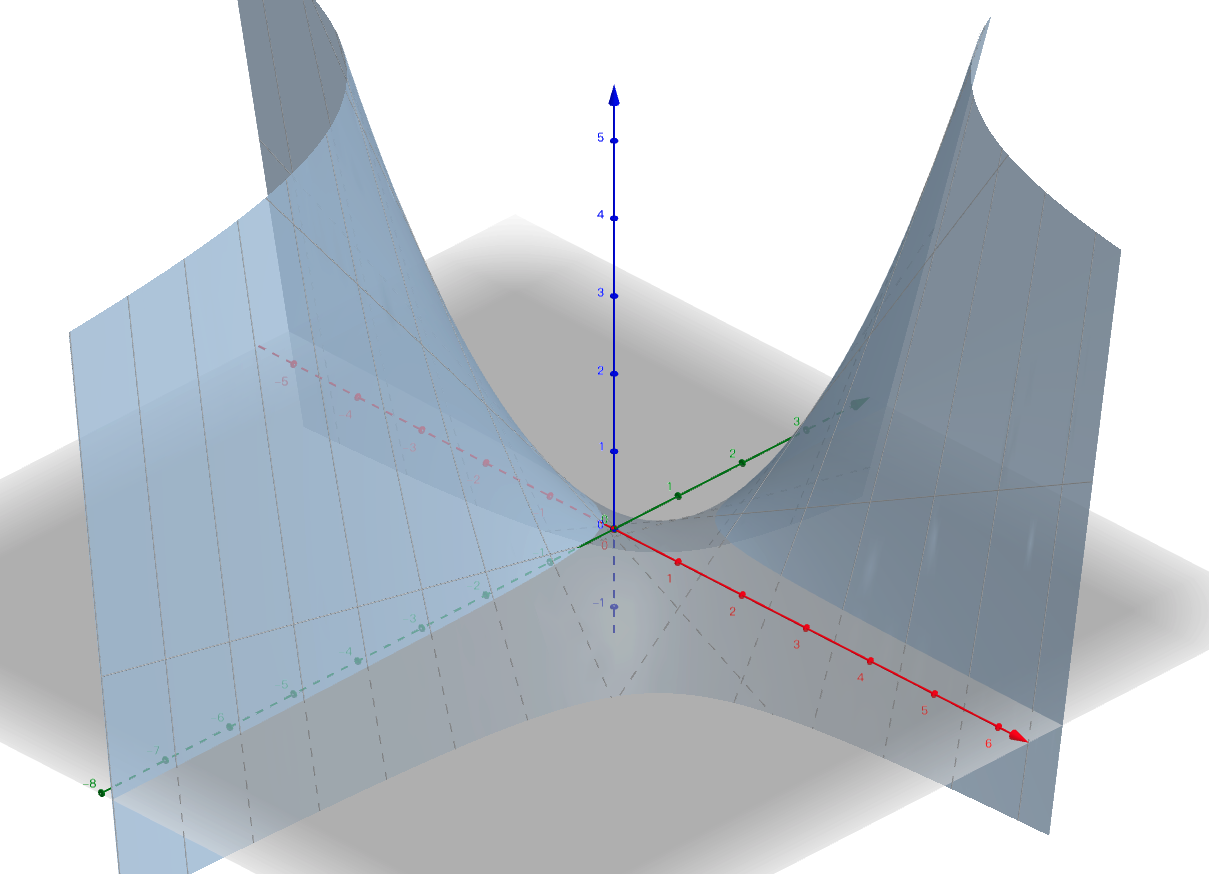}
    \end{center}
    \caption{Saddle surface}
    \label{fig:saddle}
\end{figure}
Let the objective function be :
\begin{equation}
    G^L(p_1,p_2) = p_1p_2 - (a^L p_1 + b^L p_2 + c^L)\label{con:ap1}
\end{equation}
To find the appropriate parameters, we need find the optimal value of $a^L$, $b^L$ and $c^L$ to minimize the gap function $G$, $min G^L(p_1,p_2)$. As the $p_1$ and $p_2$ are constrained by the area $[l_1, u_1] \times [l_2, u_2]$. The objective function can be reformulated as:
{\scriptsize
\begin{equation}
     \mathop{max}\limits_{a^{L},b^{L},c^{L}}F^L(p_1,p_2)=\mathop{max}\limits_{a^{L},b^{L},c^{L}}\int_{p_1 \in\left[l_1, u_1\right]} \int_{p_2 \in\left[l_2, u_2\right]} a^{L} p_1+b^{L} p_2+c^{L} 
\end{equation}}
s.t. $G^L(p_1,p_2) \geq 0 \text{ } (\forall(p_1, p_2) \in\left[l_1, u_1\right] \times\left[l_2, u_2\right])$. 

First, we need to ensure that $G^L(p_1,p_2) \geq 0$ by finding the minimum value of $G(p_1,p_2)$. The first-order partial derivatives of $G^L$ are: $\frac{\partial G^{L}}{\partial p_1}=p_2-a^{L}, \frac{\partial G^{L}}{\partial p_2}=p_1-b^{L}$.
The critical points are $\frac{\partial G^{L}}{\partial p_1} =0$ and $\frac{\partial G^{L}}{\partial p_2} =0$, which leads to $p_2 = a^L$ and $p_1 = b^L $.
Then we form the Hessian matrix to justify the behavior of the critical points:
{\footnotesize
\begin{equation}
\mathrm{H}(G^L(p_1, p_2))=\left(\begin{array}{cc}\frac{\partial^{2} G^L}{\partial p_1^2} & \frac{\partial^{2} G^L}{\partial p_1 \partial p_2} \\ \frac{\partial^{2} G^L}{\partial p_2 \partial p_1} &\frac{\partial^{2} G^L}{\partial p_2^2}\end{array}\right) = \left(\begin{array}{cc}0 & 1 \\1 & 0\end{array}\right)
\end{equation}}
Then we compute the Hessian:
{\footnotesize
\begin{equation}
\operatorname{det} \mathrm{H}(G^L(p_1, p_2))=\left(\frac{\partial^{2} G^L}{\partial p_1^{2}}\right)\left(\frac{\partial^{2} G^L}{\partial p_2^{2}}\right)-\frac{\partial^{2} G^L}{\partial p_1 \partial p_2} = -1 <0
\end{equation}}

 Thus, the critical points are saddle points, which can be illustrated in Figure \ref{fig:saddle}. Thus, the minimum value of $G^L(p_1,p_2)$ can be found on the boundary. As there exists a point $(p_1^0,p_2^0) \text{} \in [l_1,u_1] \times [l_2,u_2]$ such that $G^L(p_1^0,p_2^0) = 0$. Thereby, we need to satisfy constraints on $G^L(p_1,p_2)$ that $G^L(p_1^0,p_2^0) = 0$ and the value of $G^L$ on boundary points:  $G^L(l_1,u_1),G^L(l_1,u_2),G^L(l_2,u_1),G^L(l_2,u_2) \geq 0$.
 Equivalently, this can be formulated as:
 {\footnotesize
 \begin{equation}
\left\{\begin{array}{c}c=p_1^{0} p_2^{0}-a^{L} p_1^{0}-b^{L} p_2^{0} \\ l_{1} l_{2}-a^{L}\left(l_{1}-p_1^{0}\right)-b^{L}\left(l_{1}-p_2^{0}\right)-p_1^{0} p_2^{0} \geq 0 \\ l_{1} u_{2}-a^{L}\left(l_{1}-p_1^{0}\right)-b^{L}\left(u_{2}-p_2^{0}\right)-p_1^{0} p_2^{0} \geq 0 \\ u_{1} l_{2}-a^{L}\left(u_{1}-p_1^{0}\right)-b^{L}\left(l_{2}-p_2^{0}\right)-p_1^{0} p_2^{0} \geq 0 \\ u_{1} u_{2}-a^{L}\left(u_{1}-p_1^{0}\right)-b^{L}\left(u_{2}-p_2^{0}\right)-p_1^{0} p_2^{0} \geq 0\end{array}\right.\label{con:ap2}
\end{equation}}
 By substituting the $c^L$ in Eq. \ref{con:ap1} with equation Eq. \ref{con:ap2}, we can obtain:
 {\footnotesize
 \begin{equation}
     F^L(p_1,p_2) = \Lambda[(l_1 +u_1 -2p_1^0)a^L + (l_2 + u_2 - 2p_2)b^L + 2p_1^0p_2^0],\label{con:ap3}
 \end{equation}}
 where $\Lambda = \frac{(u_1-l_1)(u_2-l_2)}{2}$. As the minimum value can only be found on the boundary, the $p_1^0$ can be $l_1$ or $u_1$.
 \begin{itemize}
    \item If $p_1^0 = l_1$: By substituting it in Eq. \ref{con:ap2} and Eq. \ref{con:ap3} , we get:
    {\footnotesize
    \begin{equation}
    \left\{\begin{array}{c}
    a^{L} \leq \frac{u_{1} l_{2}-l_{1} p_2^{0}-b^{L}\left(l_{2}-p_2^{0}\right)}{u_{1}-l_{1}}, a^{L} \leq \frac{u_{1} u_{2}-l_{1} p_2^{0}-b^{L}\left(u_{2}-p_2^{0}\right)}{u_{1}-l_{1}}, l_{1} \leq b^{L} \leq l_{1} \Leftrightarrow b^{L}=l_{1}.\\
    F^L(p_1,p_2) = \Lambda[(u_1 -l_1)a^L + (l_2 + u_2 - 2p_2^0)b^L + 2l_1p_2^0].\end{array}\right.
    \label{con:ap4}
    \end{equation}}
    Next, replacing $b^{L}=l_{1}$, we can obtain:
    {\footnotesize $$\left\{\begin{array}{c}
    F^L(p_1,p_2) =\Lambda[(u_1 -l_1)a^L + l_1(l_2 + u_2)],\\
    \left(u_{1}-l_{1}\right)a^{L} \leq-l_{1} p_2^{0}+\min \left(u_{1} l_{2}-b^{L}\left(l_{2}-p_2^{0}\right), u_{1} u_{2}-b^{L}\left(u_{2}-p_2^{0}\right)\right)=\left(u_{1}-l_{1}\right) l_{2} . \end{array}\right.$$}
  
    As a result, we obtain $a^L \leq l_2$.
    To maximize $F^L(p_1,p_2)$, because $\Lambda \geq 0$, we adapt $a^L = l_2$ to get $F^L (p_1,p_2) = \Lambda(u_1l_2+l_1u_2)$, which is constant.
    \item If $p_1^0 = u_1$:Similarly, we can compute 
    {\footnotesize
    $$
    \left\{\begin{array}{c} a^{L} \geq \frac{l_{1} l_{2}-u_{1} p_2^{0}-b^{L}\left(l_{2}-p_2^{0}\right)}{l_{1}-u_{1}},
    \begin{array}{c}a^{L} \geq \frac{l_{1} u_{2}-u_{1} p_2^{0}-b^{L}\left(u_{2}-p_2^{0}\right)}{l_{1}-u_{1}}, u_{1} \leq b^{L} \leq u_{1} \Leftrightarrow b^{L}=u_{1}\end{array}.\\
    F^{L}(p_1,p_2)=\Lambda\left[\left(l_{1}-u_{1}\right) a^{L}+\left(l_{2}+u_{2}-2 p_2^{0}\right) b^{L}+2 u_{1} p_2^{0}\right]\end{array}\right.
    $$}
    By replacing $b^{L}=u_{1}$, we obtain
     {\footnotesize
    $$
    \left\{\begin{array}{c} F^{L}(p_1,p_2)=\Lambda\left[\left(l_{1}-u_{1}\right), a^{L}+u_{1}\left(l_{2}+u_{2}\right)\right],\\ \left(l_{1}-u_{1}\right) a^{L} \leq -u_{1} p_2^{0}+\min \left(l_{1} l_{2}-b^{L}\left(l_{2}-p_2^{0}\right), l_{1} u_{2}-b^{L}\left(u_{2}-p_2^{0}\right)\right)=\left(l_{1}-u_{1}\right) u_{2} \end{array}\right.
    $$}
    Then, we conclude that $a^L \geq u_y$, and $F^L = \Lambda(l_1u_2+u_1l_2)$
 \end{itemize}
 As the $F^L$ are the same for above two cases, which is not determined by $p_2^0$, we can choose $p_2^0 = l_2$. Thus we can obtain the parameters to compute the lower bounds: $a^{L}=l_{2}, b^{L}=l_{1}, c^{L}=-l_{1} l_{2}$
\subsection{derivation for the upper bound}
Similarly, to compute the upper bound, we can formulate the $G^U(p_1,p_2)$ as:
{\scriptsize
\begin{equation}
    G^U(p_1,p_2) = (a^U p_1 + b^U p_2 + c^U) - p_1p_2 
\end{equation}}
Thus, the objective function to compute the upper bound is:
{\footnotesize
\begin{equation}
     \mathop{min}\limits_{a^{U},b^{U},c^{U}}F^U(p_1,p_2)=\mathop{min}\limits_{a^{U},b^{U},c^{U}}\int_{p_1 \in\left[l_1, u_1\right]} \int_{p_2 \in\left[l_2, u_2\right]} a^{U} p_1+b^{U} p_2+c^{U},
\end{equation}}
and then it can be expressed as:
 {\footnotesize$$F^U(p_1,p_2) = \Lambda[(l_1 +u_1 -2p_1^0)a^U + (l_2 + u_2 - 2p_2)b^U + 2p_1^0p_2^0], \Lambda = \frac{(u_1-l_1)(u_2-l_2)}{2}$$}
 \begin{itemize}
     \item If we set $p_1^0 = l_1$:
     {\footnotesize
    $$
    \left\{\begin{array}{c}
    a^{U} \geq \frac{u_{1} l_{2}-l_{1} p_2^{0}-b^{U}\left(l_{2}-p_2^{0}\right)}{u_{1}-l_{1}}, a^{U} \geq \frac{u_{1} u_{2}-l_{1} p_2^{0}-b^{U}\left(u_{2}-p_2^{0}\right)}{u_{1}-l_{1}}, l_{1} \leq b^{U} \leq l_{1} \Leftrightarrow b^{U}=l_{1}.\\
    F^U(p_1,p_2) = \Lambda[(u_1 -l_1)a^U + (l_2 + u_2 - 2p_2^0)b^U + 2l_1p_2^0]\end{array}\right.$$} 
    After substituting $b^{U}=l_{1}$, we get 
    {\footnotesize
    $$
    \left\{\begin{array}{c} F^U(p_1,p_2) =\Lambda[(u_1 -l_1)a^U + l_1(l_2 + u_2)], \\
   \left(u_{1}-l_{1}\right) a^{U} \geq-l_{1} p_2^{0}+\max \left(u_{1} l_{2}-b^{U}\left(l_{2}-p_2^{0}\right), u_{1} u_{2}-b^{U}\left(u_{2}-p_2^{0}\right)\right) =\left(u_{1}-l_{1}\right) u_{2}\end{array}\right.$$}
    Thus, $a^U \geq u_2$. To minimize $F^U(p_1,p_2)$, we adapt $a^U = u_2$ to get $F^U (p_1,p_2) = \Lambda(l_1l_2+u_1u_2)$.
    \item If we set $p_1^0 = l_1$:
    {\footnotesize
    $$
    \left\{\begin{array}{c} a^{U} \leq \frac{l_{1} l_{2}-u_{1} p_2^{0}-b^{U}\left(l_{2}-p_2^{0}\right)}{l_{1}-u_{1}},
    a^{U} \leq \frac{l_{1} u_{2}-u_{1} p_2^{0}-b^{U}\left(u_{2}-p_2^{0}\right)}{l_{1}-u_{1}}, u_{1} \leq b^{U} \leq u_{1} \Leftrightarrow b^{U}=u_{1}\\
    F^{U}(p_1,p_2)=\Lambda\left[\left(l_{1}-u_{1}\right) a^{U}+\left(l_{2}+u_{2}-2 p_2^{0}\right) b^{U}+2 u_{1} p_2^{0}\right]\end{array}\right.$$}
    Next, we get 
    {\footnotesize
    $$
    \left\{\begin{array}{c} F^{U}(p_1,p_2)=\Lambda\left[\left(l_{1}-u_{1}\right) a^{U}+u_{1}\left(l_{2}+u_{2}\right)\right], \\
    \left(l_{1}-u_{1}\right) a^{U} \leq -u_{1} p_2^{0}+\max \left(l_{1} l_{2}-b^{U}\left(l_{2}-p_2^{0}\right), l_{1} u_{2}-b^{U}\left(u_{2}-p_2^{0}\right)\right)=\left(l_{1}-u_{1}\right) l_{2} \end{array}\right.$$}
    Thus, we can conclude that $a^U \leq l_2$, and to minimize $F^U$, we set $a^U = l_2$ to form $F^U = \Lambda(l_1l_2+u_1u_2)$.
 \end{itemize}
 As the two results for $F^U$ are the same, we choose $p_2^0 = l_2$ to obtain the final parameters for the upper bound:
 $a^{U}=u_{2}, b^{U}=l_{1}, c^{U}=-l_{1} u_{2}$.
\section{Extra Experiments}
As computing the bounds by 3DCertify is computationally unattainable, it will terminate after verifying several samples. Thus, we present the results to verify 10 samples. From the table \ref{label:resultsap1}, we can see that our method could obtain tighter bounds and significantly reduce the time.
 \begin{table*}[h]\footnotesize
\begin{center}
\vspace{-3mm}
\scalebox{0.75}{
\begin{tabular}{p{0.07\textwidth}<{\centering}|l|p{0.06\textwidth}<{\centering}|l|p{0.04\textwidth}<{\centering}||a{0.08\textwidth}<{\centering}a{0.08\textwidth}<{\centering}|p{0.08\textwidth}<{\centering}p{0.08\textwidth}<{\centering}|p{0.07\textwidth}<{\centering}p{0.07\textwidth}<{\centering}||p{0.08\textwidth}<{\centering}|p{0.08\textwidth}<{\centering}}
\hline
no. points&Pooling&Acc.& N & {$l_p$} & \multicolumn{4}{p{0.32\textwidth}<{\centering}|}{Certified Bounds} &\multicolumn{2}{p{0.14\textwidth}<{\centering}||}{Our vs.}  & \multicolumn{2}{p{0.16\textwidth}<{\centering}}{Attack}       
\\ \cline{6-13} 
 & & & &  & \multicolumn{2}{p{0.16\textwidth}<{\centering}|}{Our} & \multicolumn{2}{p{0.14\textwidth}<{\centering}|}{3DCertify} &\multicolumn{2}{p{0.16\textwidth}<{\centering}||}{3DCertify (\%)} &CW  & PGD   \\
 \cline{6-13} 
 & & & & &{ave} & {time(s)} & ave  & time(s) & ave  & time & ave  & ave
 \\ \hline
 512& {Max } &82.12&9& $l_\infty$* & 0.0052 &  58.31& 0.0030 &487.21 & +73.33& -88.03& 0.0325 & 0.0174 
\\
   & & & &$l_2$   & 0.0159  &58.78 &-& -&- & -&0.2627 & 0.3583              
\\ 
& & & &{$l_1$}  & 0.0196 &  {54.34} &-  &{-} &  & & {4.4918}   & 5.0793  
\\ \hline
1024&\multicolumn{1}{|l|}{{Average }} & 72.61&5& $l_\infty$* & 0.0141 & 50.05 &0.0107 &472.15 &+31.77 &-89.4 &0.0512 & 0.0391
\\
   & & &&$l_2$   &0.0471  &21.88 &-& -&- & -& 0.4635& 0.3877              
\\ 
& & & &{$l_1$}  & 0.0544 &  {10.08} &-  &{-} & & & {2.3798}   & 3.0146  
\\ \hline
\end{tabular}}
\caption{Average bounds and computational times on PointNet models without JANet. For the certified bounds, the higher the better. The bounds obtained by attack based method can be viewed as upper bounds.}
\vspace{-3mm}
\label{label:resultsap1}
\end{center}
\end{table*}
\end{appendices}
\section{Configuration of Various PointNet models}
Below are the specific layer configurations for 13-layer full PointNet with JANet and PointNets without JANet. 
\begin{table}
\begin{center}
\begin{tabular}{|c c c c c|}
\hline
No & Type& no. features&normalization & activation function  \\
\hline\hline
1&Conv1D&32&BatchNormalization&ReLU \\
\hline
2 & Conv1D &64 &BatchNormalization&ReLU \\
\hline
3 & Conv1D &512 &BatchNormalization&ReLU\\
\hline
4 & GlobalMaxpooling & & & \\
\hline
5 & Dense &256 &BatchNormalization&ReLU\\
\hline
6 & Dense &512 &BatchNormalization&ReLU\\
\hline
 &Reshape & & & \\
\hline
 & Multiplication & & & \\
\hline
\end{tabular}
\end{center}

\caption{Configuration for T-Net.}
\label{label:acc1}
\end{table}
\begin{table}
\begin{center}
\begin{tabular}{|c c c c c|}
\hline
No & Type& no. features&normalization & activation function   \\
\hline\hline
&T-Net& & & \\
\hline
7&Conv1D&32&BatchNormalization&ReLU \\
\hline
8 & Conv1D &64 &BatchNormalization&ReLU \\
\hline
9 & Conv1D &512 &BatchNormalization&ReLU\\
\hline
10 & GlobalMaxpooling & & & \\
\hline
11 & Dense &512 &BatchNormalization&ReLU\\
\hline
12 & Dense &256 &BatchNormalization&ReLU\\
\hline
 & Dropout & & & \\
 \hline
\end{tabular}
\end{center}

\caption{Configuration for full PointNet with JANet.}
\label{label:ac2c}
\end{table}
\begin{table}
\begin{center}
\begin{tabular}{|c c c c c|}
\hline
No & Type& no. features&normalization & activation function   \\
\hline\hline
1&Conv1D&32&BatchNormalization&ReLU \\
\hline
2&Conv1D&32&BatchNormalization&ReLU \\
\hline
3 & Conv1D &64 &BatchNormalization&ReLU \\
\hline
4 & Conv1D &512 &BatchNormalization&ReLU\\
\hline
5 & GlobalMaxpooling & & & \\
\hline
6 & Dense &512 &BatchNormalization&ReLU\\
\hline
7 & Dense &256 &BatchNormalization&ReLU\\
\hline
 & Dropout & & & \\
 \hline
\end{tabular}
\end{center}
\caption{Configuration for 7-layer PointNet without JANet.}
\label{label:acc4}
\end{table}

\begin{table}
\begin{center}
\begin{tabular}{|c c c c c|}
\hline
No & Type& no. features&normalization & activation function   \\
\hline\hline
1&Conv1D&32&BatchNormalization&ReLU \\
\hline
2&Conv1D&32&BatchNormalization&ReLU \\
\hline
3 & Conv1D &64 &BatchNormalization&ReLU \\
\hline
4 & Conv1D &64 &BatchNormalization&ReLU\\
\hline
5 & Conv1D &512 &BatchNormalization&ReLU \\
\hline
6 & Conv1D &512 &BatchNormalization&ReLU\\
\hline
7 & GlobalMaxpooling & & & \\
\hline
8 & Dense &512 &BatchNormalization&ReLU\\
\hline
9 & Dense &256 &BatchNormalization&ReLU\\
\hline
10 & Dense &256 &BatchNormalization&ReLU\\
\hline
11 & Dense &128 &BatchNormalization&ReLU\\
\hline
12 & Dense &128 &BatchNormalization&ReLU\\
\hline
 & Dropout & & & \\
 \hline
\end{tabular}
\end{center}

\caption{Configuration for 12-layer PointNet without JANet.}
\label{label:acc3}
\end{table}
\section*{Statements and Declarations}

\subsection{Funding}

This work is supported by Partnership Resource Fund on EPSRC project on Offshore Robotics for Certification of Assets (ORCA) [EP/R026173/1].

\subsection{Conflict/Competing of interest interests}
The authors declare that they have no conflict of interest

\subsection{Ethics approval $\&$ Consent to participate/publication}
Not applicable.

\subsection{Code/Data availability} 
Our code is available on \url{https://github.com/TrustAI/3DVerifier}.

\subsection{Authors' contributions}
RM contributed to the idea, algorithm, theoretical analysis, writing, and experiments. WR contributed to the idea, theoretical analysis, and writing. LM contributed to the theoretical analysis and writing. QN contributed to the idea and writing.

\bibliography{sn-bibliography}

\begin{thebibliography}{42}
\providecommand{\natexlab}[1]{#1}
\providecommand{\url}[1]{{#1}}
\providecommand{\urlprefix}{URL }
\providecommand{\doi}[1]{\url{https://doi.org/#1}}
\providecommand{\eprint}[2][]{\url{#2}}
 \bibcommenthead

\bibitem[{Aoki et~al(2019)Aoki, Goforth, Srivatsan, and Lucey}]{Aoki_2019_CVPR}
Aoki Y, Goforth H, Srivatsan RA, et~al (2019) Pointnetlk: Robust \& efficient
  point cloud registration using pointnet. { CVPR}

\bibitem[{Athalye et~al(2018)Athalye, Carlini, and
  Wagner}]{athalye2018obfuscated}
Athalye A, Carlini N, Wagner D (2018) Obfuscated gradients give a false sense
  of security: Circumventing defenses to adversarial examples. {ICML}

\bibitem[{Boopathy et~al(2019)Boopathy, Weng, Chen, Liu, and
  Daniel}]{boopathy2018cnncert}
Boopathy A, Weng TW, Chen PY, et~al (2019) {CNN-Cert: An Efficient Framework
  for Certifying Robustness of Convolutional Neural Networks}. {AAAI}

\bibitem[{Bunel et~al(2017)Bunel, Turkaslan, Torr, Kohli, and
  Kumar}]{bunel2018unified}
Bunel R, Turkaslan I, Torr PH, et~al (2017) A unified view of piecewise linear
  neural network verification. arXiv

\bibitem[{Cao et~al(2019)Cao, Xiao, Yang, Fang, Yang, Liu, and
  Li}]{cao2019adversarial}
Cao Y, Xiao C, Yang D, et~al (2019) Adversarial objects against lidar-based
  autonomous driving systems. arXiv

\bibitem[{Carlini and Wagner(2017)}]{carlini2017evaluating}
Carlini N, Wagner D (2017) Towards evaluating the robustness of neural
  networks. {SP}

\bibitem[{Chen et~al(2017)Chen, Ma, Wan, Li, and Xia}]{chen2017multiview}
Chen X, Ma H, Wan J, et~al (2017) {Multi-view 3D object detection network for
  autonomous driving}. {CVPR}

\bibitem[{Chen et~al(2021)Chen, Jiang, Zhu, Wang, and Yun}]{chen2021individual}
Chen X, Jiang K, Zhu Y, et~al (2021) Individual tree crown segmentation
  directly from uav-borne lidar data using the pointnet of deep learning.
  Forests

\bibitem[{Dvijotham et~al(2018)Dvijotham, Stanforth, Gowal, Mann, and
  Kohli}]{krishnamurthy2018dual}
Dvijotham K, Stanforth R, Gowal S, et~al (2018) A dual approach to scalable
  verification of deep networks. UAI

\bibitem[{Gehr et~al(2018)Gehr, Mirman, Drachsler-Cohen, Tsankov, Chaudhuri,
  and Vechev}]{gehr2018ai2}
Gehr T, Mirman M, Drachsler-Cohen D, et~al (2018) {Ai2: Safety and robustness
  certification of neural networks with abstract interpretation}. {SP}

\bibitem[{Goodfellow et~al(2014)Goodfellow, Shlens, and
  Szegedy}]{goodfellow2015explaining}
Goodfellow IJ, Shlens J, Szegedy C (2014) Explaining and harnessing adversarial
  examples. arXiv

\bibitem[{Jin et~al(2020)Jin, Yi, Zhang, Zhang, Schewe, and
  Huang}]{jin2020does}
Jin G, Yi X, Zhang L, et~al (2020) How does weight correlation affect
  generalisation ability of deep neural networks? NeurIPS

\bibitem[{Jin et~al(2022)Jin, Yi, Huang, Schewe, and Huang}]{jin2022enhancing}
Jin G, Yi X, Huang W, et~al (2022) Enhancing adversarial training with
  second-order statistics of weights. In: CVPR

\bibitem[{Katz et~al(2017)Katz, Barrett, Dill, Julian, and
  Kochenderfer}]{katz2017reluplex}
Katz G, Barrett C, Dill DL, et~al (2017) {Reluplex: An efficient SMT solver for
  verifying deep neural networks}. {ICCAV}

\bibitem[{Kurakin et~al(2016)Kurakin, Goodfellow, and
  Bengio}]{kurakin2016adversarial}
Kurakin A, Goodfellow IJ, Bengio S (2016) Adversarial examples in the physical
  world. arXiv

\bibitem[{Lee et~al(2020)Lee, Chen, Yan, Urtasun, and Yumer}]{lee2020shapeadv}
Lee K, Chen Z, Yan X, et~al (2020) {ShapeAdv: Generating Shape-Aware
  Adversarial 3D Point Clouds}. arXiv

\bibitem[{Liang et~al(2018)Liang, Yang, Wang, and Urtasun}]{liang2018deep}
Liang M, Yang B, Wang S, et~al (2018) Deep continuous fusion for multi-sensor
  3d object detection. {ECCV}

\bibitem[{Liu et~al(2019)Liu, Yu, and Su}]{liu2019extending}
Liu D, Yu R, Su H (2019) {Extending adversarial attacks and defenses to deep 3D
  point cloud classifiers}. {ICIP}

\bibitem[{Lorenz et~al(2021)Lorenz, Ruoss, Balunovi{\'c}, Singh, and
  Vechev}]{lorenz2021robustness}
Lorenz T, Ruoss A, Balunovi{\'c} M, et~al (2021) Robustness certification for
  point cloud models. arXiv

\bibitem[{Paigwar et~al(2019)Paigwar, Erkent, Wolf, and
  Laugier}]{paigwar2019attentional}
Paigwar A, Erkent O, Wolf C, et~al (2019) Attentional pointnet for 3d-object
  detection in point clouds. {CVPR Workshops}

\bibitem[{Qi et~al(2017{\natexlab{a}})Qi, Su, Mo, and Guibas}]{qi2017pointnet}
Qi CR, Su H, Mo K, et~al (2017{\natexlab{a}}) {PointNet: Deep Learning on Point
  Sets for 3D Classification and Segmentation}. {CVPR}

\bibitem[{Qi et~al(2017{\natexlab{b}})Qi, Yi, Su, and
  Guibas}]{qi2017pointnet++}
Qi CR, Yi L, Su H, et~al (2017{\natexlab{b}}) {PointNet++: Deep Hierarchical
  Feature Learning on Point Sets in a Metric Space}. arXiv

\bibitem[{Salman et~al(2019)Salman, Yang, Zhang, Hsieh, and
  Zhang}]{salman2020convex}
Salman H, Yang G, Zhang H, et~al (2019) A convex relaxation barrier to tight
  robustness verification of neural networks. arXiv

\bibitem[{Shi et~al(2020)Shi, Zhang, Chang, Huang, and
  Hsieh}]{Shi2020Robustness}
Shi Z, Zhang H, Chang KW, et~al (2020) {Robustness Verification for
  Transformers}. {ICLR}

\bibitem[{Singh et~al(2019)Singh, Gehr, P{\"u}schel, and
  Vechev}]{singh2019abstract}
Singh G, Gehr T, P{\"u}schel M, et~al (2019) An abstract domain for certifying
  neural networks. {ACM PL}

\bibitem[{Stets et~al(2017)Stets, Sun, Corning, and Greenwald}]{vr3d}
Stets JD, Sun Y, Corning W, et~al (2017) {Visualization and Labeling of Point
  Clouds in Virtual Reality}. SIGGRAPH Asia

\bibitem[{Sun et~al(2020)Sun, Koenig, Cao, Chen, and Mao}]{sun2021adversarial}
Sun J, Koenig K, Cao Y, et~al (2020) {On Adversarial Robustness of 3D Point
  Cloud Classification under Adaptive Attacks}. arXiv

\bibitem[{Szegedy et~al(2014)Szegedy, Zaremba, Sutskever, Bruna, Erhan,
  Goodfellow, and Fergus}]{szegedy2014intriguing}
Szegedy C, Zaremba W, Sutskever I, et~al (2014) Intriguing properties of neural
  networks. {ICLR}

\bibitem[{Tjeng et~al(2018)Tjeng, Xiao, and Tedrake}]{tjeng2018evaluating}
Tjeng V, Xiao KY, Tedrake R (2018) Evaluating robustness of neural networks
  with mixed integer programming. {ICLR}

\bibitem[{Tramer et~al(2020)Tramer, Carlini, Brendel, and
  Madry}]{tramer2020adaptive}
Tramer F, Carlini N, Brendel W, et~al (2020) On adaptive attacks to adversarial
  example defenses. arXiv

\bibitem[{Varley et~al(2017)Varley, DeChant, Richardson, Ruales, and
  Allen}]{varley2017shape}
Varley J, DeChant C, Richardson A, et~al (2017) Shape completion enabled
  robotic grasping. {IROS}

\bibitem[{Wang et~al(2018)Wang, Pei, Whitehouse, Yang, and
  Jana}]{wang2018efficient}
Wang S, Pei K, Whitehouse J, et~al (2018) Efficient formal safety analysis of
  neural networks. arXiv

\bibitem[{Weng et~al(2018)Weng, Zhang, Chen, Song, Hsieh, Daniel, Boning, and
  Dhillon}]{weng2018fast}
Weng L, Zhang H, Chen H, et~al (2018) {Towards Fast Computation of Certified
  Robustness for ReLU Networks}. {ICML}

\bibitem[{Wicker and Kwiatkowska(2019)}]{wicker2019robustness}
Wicker M, Kwiatkowska M (2019) {Robustness of 3D deep learning in an
  adversarial setting}. {CVPR}

\bibitem[{Wong and Kolter(2018)}]{wong2018provable}
Wong E, Kolter Z (2018) Provable defenses against adversarial examples via the
  convex outer adversarial polytope. {ICML}

\bibitem[{Wu et~al(2015)Wu, Song, Khosla, Yu, Zhang, Tang, and Xiao}]{wu20153d}
Wu Z, Song S, Khosla A, et~al (2015) {3D ShapeNets: A Deep Representation for
  Volumetric Shapes}. {CVPR}

\bibitem[{Xiang et~al(2019)Xiang, Qi, and Li}]{xiang2019generating}
Xiang C, Qi CR, Li B (2019) {Generating 3D adversarial point clouds}. {CVPR}

\bibitem[{Zhang et~al(2018)Zhang, Weng, Chen, Hsieh, and
  Daniel}]{zhang2018efficient}
Zhang H, Weng TW, Chen PY, et~al (2018) Efficient neural network robustness
  certification with general activation functions. {ANIPS}

\bibitem[{Zhang et~al(2019{\natexlab{a}})Zhang, Yang, Fang, Ni, Liu, and
  Tian}]{yang2021adversarial}
Zhang Q, Yang J, Fang R, et~al (2019{\natexlab{a}}) Adversarial attack and
  defense on point sets. arXiv

\bibitem[{Zhang et~al(2019{\natexlab{b}})Zhang, Liang, Salem, and
  Jacobs}]{zhang2019defense}
Zhang Y, Liang G, Salem T, et~al (2019{\natexlab{b}}) Defense-pointnet:
  Protecting pointnet against adversarial attacks. {ICBD}

\bibitem[{Zhao et~al(2020)Zhao, Wu, Chen, and Lim}]{zhao2020isometry}
Zhao Y, Wu Y, Chen C, et~al (2020) {On isometry robustness of deep 3D point
  cloud models under adversarial attacks}. {CVPR}

\bibitem[{Zhou et~al(2019)Zhou, Chen, Zhang, Fang, Zhou, and Yu}]{zhou2019dup}
Zhou H, Chen K, Zhang W, et~al (2019) {Dup-net: Denoiser and upsampler network
  for 3D adversarial point clouds defense}. {ICCV}

\end{thebibliography}
\end{document}


\title[3DVerifier: Efficient Robustness Verification for 3D Point Cloud Models]{{\bf ~~~~Supplementary Document}\\3DVerifier: Efficient Robustness Verification for 3D Point Cloud Models} 




\maketitle
\appendix
\section{The Power of JANet}

We perform ablation study to show the importance of JANet in the point clouds classification task. We choose three models to train, and record the number of trainable parameters for each model. The final results of training accuracy are demonstrated in Table \ref{label:accJAnet}. The specific layer configurations for these models are presented in Appendix C. As the PointNet without JANet can be regarded as a general CNN model, while JANet is a complex architecture that contains T-Net and multiplication layer, to interpret the effect of JANet, we examine two models for the PointNet without JANet: 1) 7-layer model, which abandons the JANet in the 13-layer full PointNet; 2) 13-layer model, which intuitively adds the convolution and dense layers in T-Net to the 7-layer model. From the results, we can see that the PointNet with JANet can improve the training accuracy significantly comparing with PointNet without JANet when they employ the same number of layers. Therefore, JANet plays an important role to improve the performance of point clouds models.

\begin{table}
\begin{center}
\caption{Training accuracy of different structures of point cloud models.}
\label{label:accJAnet}
\begin{tabular}{|l|c|c|c|}
\hline
Models & N&no. trainable parameters & accuracy  \\
\hline\hline
PointNet without JANet &7& 443656 & 72.94\% \\
\hline
PointNet without JANet & 13 &822472 &74.07\% \\
\hline
Full PointNet with JANet & 13 & 645425 & 81.31\%\\
\hline
\end{tabular}
\end{center}
\end{table}

\section{Configuration of PointNet models}

Below are the specific layer configurations for 13-layer full PointNet with JANet and PointNets without JANet. 
\begin{table}[H]
\begin{center}
\vspace{-3mm}
\scalebox{0.75}{
\begin{tabular}{|c c c c c|}
\hline
No & Type& no. features&normalization & activation function  \\
\hline\hline
1&Conv1D&32&BatchNormalization&ReLU \\
\hline
2 & Conv1D &64 &BatchNormalization&ReLU \\
\hline
3 & Conv1D &512 &BatchNormalization&ReLU\\
\hline
4 & GlobalMaxpooling & & & \\
\hline
5 & Dense &256 &BatchNormalization&ReLU\\
\hline
6 & Dense &512 &BatchNormalization&ReLU\\
\hline
 &Reshape & & & \\
\hline
 & Multiplication & & & \\
\hline
\end{tabular}}
\end{center}
\caption{Configuration for T-Net.}
\label{label:con1}
\vspace{-3mm}
\end{table}

\begin{table}[H]
\begin{center}
\vspace{-3mm}
\scalebox{0.75}{
\begin{tabular}{|c c c c c|}
\hline
No & Type& no. features&normalization & activation function   \\
\hline\hline
&T-Net& & & \\
\hline
7&Conv1D&32&BatchNormalization&ReLU \\
\hline
8 & Conv1D &64 &BatchNormalization&ReLU \\
\hline
9 & Conv1D &512 &BatchNormalization&ReLU\\
\hline
10 & GlobalMaxpooling & & & \\
\hline
11 & Dense &512 &BatchNormalization&ReLU\\
\hline
12 & Dense &256 &BatchNormalization&ReLU\\
\hline
 & Dropout & & & \\
 \hline
\end{tabular}}
\end{center}
\caption{Configuration for full PointNet with JANet.}
\label{label:con2}
\vspace{-3mm}
\end{table}

\begin{table}[H]
\begin{center}
\vspace{-3mm}
\scalebox{0.8}{
\begin{tabular}{|c c c c c|}
\hline
No & Type& no. features&normalization & activation function   \\
\hline\hline
1&Conv1D&32&BatchNormalization&ReLU \\
\hline
2&Conv1D&32&BatchNormalization&ReLU \\
\hline
3 & Conv1D &64 &BatchNormalization&ReLU \\
\hline
4 & Conv1D &512 &BatchNormalization&ReLU\\
\hline
5 & GlobalMaxpooling & & & \\
\hline
6 & Dense &512 &BatchNormalization&ReLU\\
\hline
7 & Dense &256 &BatchNormalization&ReLU\\
\hline
 & Dropout & & & \\
 \hline
\end{tabular}}
\end{center}
\caption{Configuration for 7-layer PointNet without JANet.}
\label{label:con3}
\vspace{-4mm}
\end{table}

\begin{table}[H]
\begin{center}
\scalebox{0.9}{
\begin{tabular}{|c c c c c|}
\hline
No & Type& no. features&normalization & activation function   \\
\hline\hline
1&Conv1D&32&BatchNormalization&ReLU \\
\hline
2&Conv1D&32&BatchNormalization&ReLU \\
\hline
3 & Conv1D &64 &BatchNormalization&ReLU \\
\hline
4 & Conv1D &64 &BatchNormalization&ReLU\\
\hline
5 & Conv1D &512 &BatchNormalization&ReLU \\
\hline
6 & Conv1D &512 &BatchNormalization&ReLU\\
\hline
7 & GlobalMaxpooling & & & \\
\hline
8 & Dense &512 &BatchNormalization&ReLU\\
\hline
9 & Dense &256 &BatchNormalization&ReLU\\
\hline
10 & Dense &256 &BatchNormalization&ReLU\\
\hline
11 & Dense &128 &BatchNormalization&ReLU\\
\hline
12 & Dense &128 &BatchNormalization&ReLU\\
\hline
 & Dropout & & & \\
 \hline
\end{tabular}}
\end{center}
\caption{Configuration for 12-layer PointNet without JANet.}
\label{label:con4}
\end{table}
\section{Derivation of the linear functions to bound the multiplication layer}
The mathematical proof to choose the optimal parameters of the linear functions to bound the multiplication layer is evaluated in \cite{Shi2020Robustness}. Here we will present how to derive the optimisation problem based on the derivation in \cite{Shi2020Robustness}.

The goal of the linear relaxation is to bound the multiplication on two inputs, $p_1p_2$, which can be represented by the following constraints:
\begin{equation}
    a^L p_1 + b^L p_2 + c^L \leq p_1p_2 \leq a^Up_1 + b^Up_2 + c^U
\end{equation}
As we have already obtained the bounds of $p_1$ and $p_2$, given the bounds of $p_1$ is $[l_1, u_1]$ and for $p_2$ is $[l_2, u_2]$, we can choose the appropriate parameters of $a^L$, $a^U$, $b^L$, $b^U$, $c^L$ and $c^U$ according to the computed bounds.

\subsection{derivation for the lower bound}
Let the objective function be :
\begin{equation}
    G^L(p_1,p_2) = p_1p_2 - (a^L p_1 + b^L p_2 + c^L)
\end{equation}
Thus, to find the appropriate parameters, we need find the optimal value of $a^L$, $b^L$ and $c^L$ to minimize the gap function $G$, $min G^L(p_1,p_2)$. As the $p_1$ and $p_2$ are constrained by the area $[l_1, u_1] \times [l_2, u_2]$. Thus, we formulate the objective function as :
\begin{equation}
     \mathop{max}\limits_{a^{L},b^{L},c^{L}}F^L(p_1,p_2)=\mathop{max}\limits_{a^{L},b^{L},c^{L}}\int_{p_1 \in\left[l_1, u_1\right]} \int_{p_2 \in\left[l_2, u_2\right]} a^{L} p_1+b^{L} p_2+c^{L} 
\end{equation}
s.t. $G^L(p_1,p_2) \geq 0 \text{ } (\forall(p_1, p_2) \in\left[l_1, u_1\right] \times\left[l_2, u_2\right])$. 
First, we need to ensure that $G^L(p_1,p_2) \geq 0$ by finding the minimum value of $G(p_1,p_2)$.
The first-order partial derivatives of $G^L$ are:
\begin{equation}
\begin{array}{l}\frac{\partial G^{L}}{\partial p_1}=p_2-a^{L} \\ \frac{\partial G^{L}}{\partial p_2}=p_1-b^{L}\end{array}
\end{equation}
The critical points are $\frac{\partial G^{L}}{\partial p_1} =0$ and $\frac{\partial G^{L}}{\partial p_2} =0$, which leads to $p_2 = a^L$ and $p_1 = b^L $.
Then we form the Hessian matrix to justify the behavior of the critical points:\begin{equation}
\mathrm{H}(G^L(p_1, p_2))=\left(\begin{array}{cc}\frac{\partial^{2} G^L}{\partial p_1^2} & \frac{\partial^{2} G^L}{\partial p_1 \partial p_2} \\ \frac{\partial^{2} G^L}{\partial p_2 \partial p_1} &\frac{\partial^{2} G^L}{\partial p_2^2}\end{array}\right) = \left(\begin{array}{cc}0 & 1 \\1 & 0\end{array}\right)
\end{equation}
Then we compute the Hessian:
\begin{equation}
\operatorname{det} \mathrm{H}(G^L(p_1, p_2))=\left(\frac{\partial^{2} G^L}{\partial p_1^{2}}\right)\left(\frac{\partial^{2} G^L}{\partial p_2^{2}}\right)-\frac{\partial^{2} G^L}{\partial p_1 \partial p_2} = -1 <0
\end{equation}
\begin{figure}
    \begin{center}
    \includegraphics[scale=0.1]{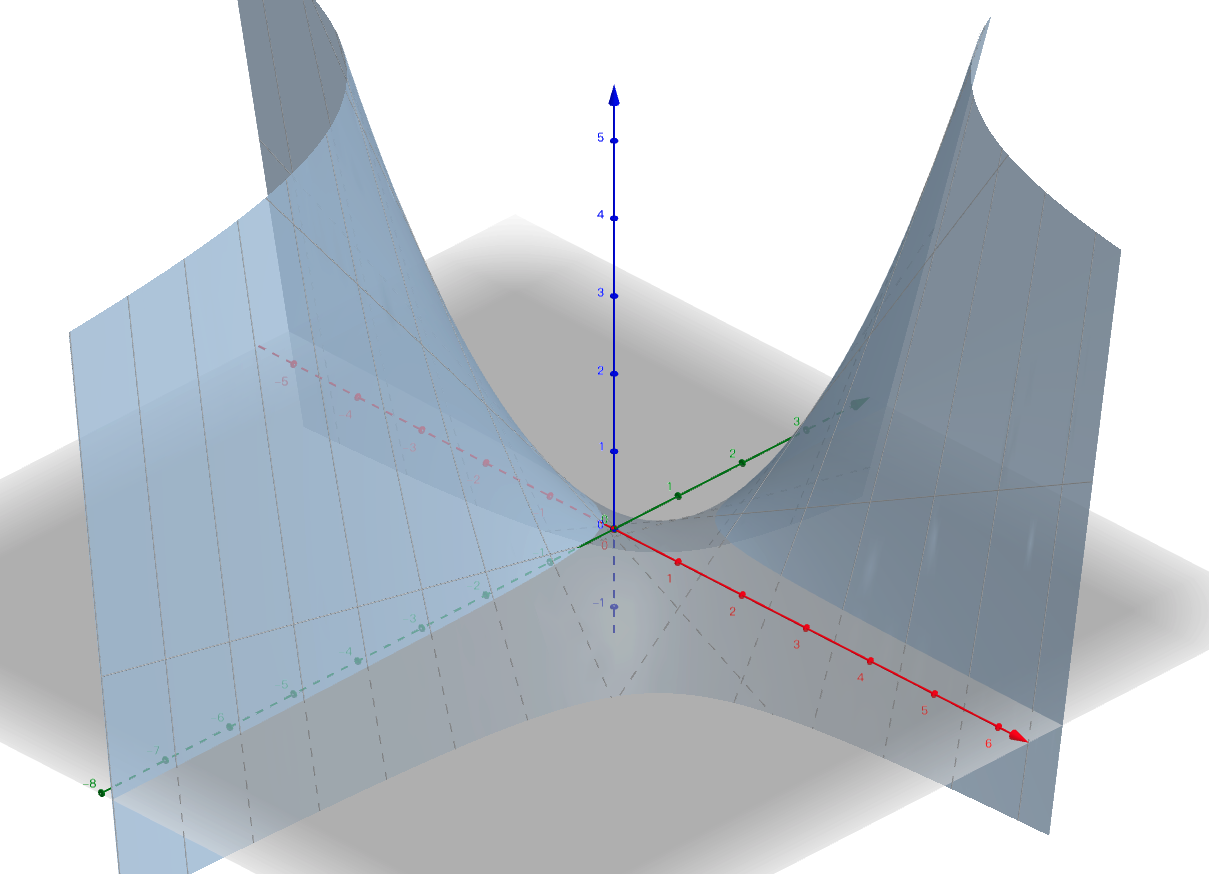}
    \end{center}
    \caption{Saddle surface}
    \label{fig:saddle}
\end{figure}
 Thus, the critical points are saddle points, which can be illustrated in Figure \ref{fig:saddle}. Thus, the minimum value of $G^L(p_1,p_2)$ can be found on the boundary. As there exists a point $(p_1^0,p_2^0) \text{} \in [l_1,u_1] \times [l_2,u_2]$ such that $G^L(p_1^0,p_2^0) = 0$. Thereby, we need to satisfy constraints on $G^L(p_1,p_2)$ that $G^L(p_1^0,p_2^0) = 0$ and the value of $G^L$ on boundary points:  $G^L(l_1,u_1),G^L(l_1,u_2),G^L(l_2,u_1),G^L(l_2,u_2) \geq 0$.
 Equivalently, this can be formulated as:
 \begin{equation}
\left\{\begin{array}{c}c=p_1^{0} p_2^{0}-a^{L} p_1^{0}-b^{L} p_2^{0} \\ l_{1} l_{2}-a^{L}\left(l_{1}-p_1^{0}\right)-b^{L}\left(l_{1}-p_2^{0}\right)-p_1^{0} p_2^{0} \geq 0 \\ l_{1} u_{2}-a^{L}\left(l_{1}-p_1^{0}\right)-b^{L}\left(u_{2}-p_2^{0}\right)-p_1^{0} p_2^{0} \geq 0 \\ u_{1} l_{2}-a^{L}\left(u_{1}-p_1^{0}\right)-b^{L}\left(l_{2}-p_2^{0}\right)-p_1^{0} p_2^{0} \geq 0 \\ u_{1} u_{2}-a^{L}\left(u_{1}-p_1^{0}\right)-b^{L}\left(u_{2}-p_2^{0}\right)-p_1^{0} p_2^{0} \geq 0\end{array}\right.
\end{equation}
 By substituting the $c^L$ in the equation (26) with equation (30), we can obtain:
 $$F^L(p_1,p_2) = \Lambda[(l_1 +u_1 -2p_1^0)a^L + (l_2 + u_2 - 2p_2)b^L + 2p_1^0p_2^0],$$
 where $\Lambda = \frac{(u_1-l_1)(u_2-l_2)}{2}$.
 
 As the minimum value can only be found on the boundary, the minimum value $(x_0,y_0)$ can only be chosen from the border. Thus, the $p_1^0$ can be $l_1$ or $u_1$.
 \begin{itemize}
    \item If $p_1^0 = l_1$:
     
     By substituting it in equation (8), we get:
    \begin{equation}
    \left\{\begin{array}{c}a^{L} \leq \frac{u_{1} l_{2}-l_{1} p_2^{0}-b^{L}\left(l_{2}-p_2^{0}\right)}{u_{1}-l_{1}} \\ a^{L} \leq \frac{u_{1} u_{2}-l_{1} p_2^{0}-b^{L}\left(u_{2}-p_2^{0}\right)}{u_{1}-l_{1}} \\ l_{1} \leq b^{L} \leq l_{1} \Leftrightarrow b^{L}=l_{1}\end{array}\right.
    \end{equation}
    And also:
    \begin{equation}
    F^L(p_1,p_2) = \Lambda[(u_1 -l_1)a^L + (l_2 + u_2 - 2p_2^0)b^L + 2l_1p_2^0].
    \end{equation}
 
    Next, replacing $b^{L}=l_{1}$ in equation (32), we can obtain:
    $$F^L(p_1,p_2) =\Lambda[(u_1 -l_1)a^L + l_1(l_2 + u_2)], $$
    and
    \begin{equation}
    \begin{aligned}\left(u_{1}-l_{1}\right) a^{L} & \leq-l_{1} p_2^{0}+\min \left(u_{1} l_{2}-b^{L}\left(l_{2}-p_2^{0}\right), u_{1} u_{2}-b^{L}\left(u_{2}-p_2^{0}\right)\right) \\ &=\left(u_{1}-l_{1}\right) \min \left(l_{2}, u_{2}\right) \\ &=\left(u_{1}-l_{1}\right) l_{2} . \end{aligned}
    \end{equation}
    As a result, we obtain $a^L \leq l_2$.
    To maximize $F^L(p_1,p_2)$, because $\Lambda \geq 0$, we adapt $a^L = l_2$ to get $F^L (p_1,p_2) = \Lambda(u_1l_2+l_1u_2)$, which is constant.
    \item If we choose $p_1^0 = u_1$:

    Similarly, we can compute 
    \begin{equation}
    \left\{\begin{array}{c} a^{L} \geq \frac{l_{1} l_{2}-u_{1} p_2^{0}-b^{L}\left(l_{2}-p_2^{0}\right)}{l_{1}-u_{1}}\\
    \begin{array}{c}a^{L} \geq \frac{l_{1} u_{2}-u_{1} p_2^{0}-b^{L}\left(u_{2}-p_2^{0}\right)}{l_{1}-u_{1}} \\ u_{1} \leq b^{L} \leq u_{1} \Leftrightarrow b^{L}=u_{1}\end{array}
    \end{array}\right.
    \end{equation}
    and
    $$F^{L}(p_1,p_2)=\Lambda\left[\left(l_{1}-u_{1}\right) a^{L}+\left(l_{2}+u_{2}-2 p_2^{0}\right) b^{L}+2 u_{1} p_2^{0}\right]$$
    By replacing $b^{L}=u_{1}$, we obtain
    $$
    \begin{aligned} & F^{L}(p_1,p_2)=\Lambda\left[\left(l_{1}-u_{1}\right) a^{L}+u_{1}\left(l_{2}+u_{2}\right)\right] \\\left(l_{1}-u_{1}\right) a^{L} \leq &-u_{1} p_2^{0}+\min \left(l_{1} l_{2}-b^{L}\left(l_{2}-p_2^{0}\right), l_{1} u_{2}-b^{L}\left(u_{2}-p_2^{0}\right)\right) \\=&\left(l_{1}-u_{1}\right) \max \left(l_{2}, u_{2}\right) \\=&\left(l_{1}-u_{1}\right) u_{2} \end{aligned}
    $$
    Then, we can conclude that $a^L \geq u_y$, and $F^L = \Lambda(l_1u_2+u_1l_2)$
 \end{itemize}
 As the $F^L$ are the same for above two cases, and it is not determined with $p_2^0$, we will choose $p_2^0 = l_2$. Thus we can obtain the parameters to compute the lower bounds:
 $$\left\{\begin{array}{l}a^{L}=l_{2} \\ b^{L}=l_{1} \\ c^{L}=-l_{1} l_{2}\end{array}\right.$$
\subsection{derivation for the upper bound}
Similarly, to compute the upper bound, we can formulate the $G^U(p_1,p_2)$ as:
\begin{equation}
    G^U(p_1,p_2) = (a^U p_1 + b^U p_2 + c^U) - p_1p_2 
\end{equation}
Thus, the objective function to compute the upper bound is:
\begin{equation}
     \mathop{min}\limits_{a^{U},b^{U},c^{U}}F^U(p_1,p_2)=\mathop{min}\limits_{a^{U},b^{U},c^{U}}\int_{p_1 \in\left[l_1, u_1\right]} \int_{p_2 \in\left[l_2, u_2\right]} a^{U} p_1+b^{U} p_2+c^{U},
\end{equation}
and then it can be expressed as:
 $$F^U(p_1,p_2) = \Lambda[(l_1 +u_1 -2p_1^0)a^U + (l_2 + u_2 - 2p_2)b^U + 2p_1^0p_2^0],$$
 where $\Lambda = \frac{(u_1-l_1)(u_2-l_2)}{2}$.
 \begin{itemize}
     \item If we set $p_1^0 = l_1$:
     We obtain the constraints as:
    \begin{equation}
    \left\{\begin{array}{c}a^{U} \geq \frac{u_{1} l_{2}-l_{1} p_2^{0}-b^{U}\left(l_{2}-p_2^{0}\right)}{u_{1}-l_{1}} \\ a^{U} \geq \frac{u_{1} u_{2}-l_{1} p_2^{0}-b^{U}\left(u_{2}-p_2^{0}\right)}{u_{1}-l_{1}} \\ l_{1} \leq b^{U} \leq l_{1} \Leftrightarrow b^{U}=l_{1}\end{array}\right.
    \end{equation}
    and
    \begin{equation}
    F^U(p_1,p_2) = \Lambda[(u_1 -l_1)a^U + (l_2 + u_2 - 2p_2^0)b^U + 2l_1p_2^0].
    \end{equation} 
    After substituting $b^{U}=l_{1}$,
    $$F^U(p_1,p_2) =\Lambda[(u_1 -l_1)a^U + l_1(l_2 + u_2)], $$
    and
    \begin{equation}
    \begin{aligned}\left(u_{1}-l_{1}\right) a^{U} & \geq-l_{1} p_2^{0}+\max \left(u_{1} l_{2}-b^{U}\left(l_{2}-p_2^{0}\right), u_{1} u_{2}-b^{U}\left(u_{2}-p_2^{0}\right)\right) \\ &=\left(u_{1}-l_{1}\right) \max \left(l_{2}, u_{2}\right) \\ &=\left(u_{1}-l_{1}\right) u_{2} . \end{aligned}
    \end{equation}
     $$a^U \geq u_2.$$
    To minimize $F^U(p_1,p_2)$, we adapt $a^U = u_2$ to get $F^U (p_1,p_2) = \Lambda(l_1l_2+u_1u_2)$.
    \item If we set $p_1^0 = l_1$:
    \begin{equation}
    \left\{\begin{array}{c} a^{U} \leq \frac{l_{1} l_{2}-u_{1} p_2^{0}-b^{U}\left(l_{2}-p_2^{0}\right)}{l_{1}-u_{1}}\\
    \begin{array}{c}a^{U} \leq \frac{l_{1} u_{2}-u_{1} p_2^{0}-b^{U}\left(u_{2}-p_2^{0}\right)}{l_{1}-u_{1}} \\ u_{1} \leq b^{U} \leq u_{1} \Leftrightarrow b^{U}=u_{1}\end{array}
    \end{array}\right.
    \end{equation}
    and
    $$F^{U}(p_1,p_2)=\Lambda\left[\left(l_{1}-u_{1}\right) a^{U}+\left(l_{2}+u_{2}-2 p_2^{0}\right) b^{U}+2 u_{1} p_2^{0}\right]$$
    Next,
    $$
    \begin{aligned} & F^{U}(p_1,p_2)=\Lambda\left[\left(l_{1}-u_{1}\right) a^{U}+u_{1}\left(l_{2}+u_{2}\right)\right] \\\left(l_{1}-u_{1}\right) a^{U} \leq &-u_{1} p_2^{0}+\max \left(l_{1} l_{2}-b^{U}\left(l_{2}-p_2^{0}\right), l_{1} u_{2}-b^{U}\left(u_{2}-p_2^{0}\right)\right) \\=&\left(l_{1}-u_{1}\right) \min \left(l_{2}, u_{2}\right) \\=&\left(l_{1}-u_{1}\right) l_{2} \end{aligned}
    $$
    Thus, we can conclude that $a^U \leq l_2$, and to minimize $F^U$, we set$a^U = l_2$ to form $F^U = \Lambda(l_1l_2+u_1u_2)$.
 \end{itemize}
 As the two results for $F^U$ are the same, we choose $p_2^0 = l_2$ to obtain the final parameters for the upper bound:
 $$\left\{\begin{array}{l}a^{U}=u_{2} \\ b^{U}=l_{1} \\ c^{U}=-l_{1} u_{2}\end{array}\right.$$

\bibliographystyle{splncs04}
\bibliography{ref}